\documentclass[10pt,twocolumn,letterpaper]{article}

\usepackage{iccv}
\usepackage{times}
\usepackage{epsfig}
\usepackage{graphicx}
\usepackage{amsmath}
\usepackage{amssymb}

\usepackage[pagebackref=true,breaklinks=true,letterpaper=true,colorlinks,bookmarks=false]{hyperref}

 \iccvfinalcopy % *** Uncomment this line for the final submission

\def\iccvPaperID{102} % *** Enter the ICCV Paper ID here

\ificcvfinal\fi
\begin{document}

%%%%%%%%% TITLE
\title{Amulet: Aggregating Multi-level Convolutional Features \\ for Salient Object Detection}

\author{Pingping Zhang$^{\dagger}$\quad Dong Wang$^{\dagger}$\quad Huchuan Lu$^{\dagger}$\thanks{Prof.Lu is the corresponding author.}\quad Hongyu Wang$^{\dagger}$\quad Xiang Ruan$^{\ddagger}$\\
%School of Information and Communication Engineering\\
$^{\dagger}$Dalian University of Technology, China\quad\quad $^{\ddagger}$Tiwaki Co.Ltd\\
%{\tt\small \{jssxzhpp,wdice,lhchuan,whyu\}@gmail.com\quad\quad ruanxiang@tiwaki.com}
{\tt\small jssxzhpp@mail.dlut.edu.cn\quad \{wdice,lhchuan,whyu\}@dlut.edu.cn\quad ruanxiang@tiwaki.com}
%\and
%Xiang Ruan\\
%Tiwaki Co.Ltd, Japan\\
%%Kusatsu, Shiga, Japan\\
%{\tt\small ruanxiang@tiwaki.com}
}

\maketitle
\thispagestyle{empty}

%%%%%%%%% ABSTRACT
\begin{abstract}
Fully convolutional neural networks (FCNs) have shown outstanding performance in many dense
labeling problems.
%lots of pixel-wise vision tasks, e.g. semantic segmentation, contour detection and scene parsing.
%
One key pillar of these successes is mining relevant information from features in convolutional layers.
However, how to better aggregate multi-level convolutional feature maps for salient object detection is underexplored.
In this work, we present \textbf{Amulet}, a generic aggregating multi-level convolutional feature framework for salient object detection.
Our framework first integrates multi-level feature maps into multiple resolutions, which simultaneously incorporate coarse semantics and fine details.
Then it adaptively learns to combine these feature maps at each resolution and predict saliency maps with the combined features. %adaptively
Finally, the predicted results are efficiently fused to generate the final saliency map.
In addition, to achieve accurate boundary inference and semantic enhancement, edge-aware feature maps in low-level layers and the predicted results of low resolution features are recursively embedded into the learning framework.
By aggregating multi-level convolutional features in this efficient and flexible manner, the proposed saliency model provides accurate salient object labeling.
Comprehensive experiments demonstrate that our method performs favorably against state-of-the-art approaches in terms of near all compared evaluation metrics.
\end{abstract}

%%%%%%%%% BODY TEXT
\section{Introduction}
Salient object detection, which aims to identify the most conspicuous objects or regions in an image, has received considerable amount of attention in recent years.
%image caption~\cite{Vinyals2014Show,Xu2015Show,Fang2015From,densecap},~\cite{mahadevan2009saliency},
As a pre-processing step in computer vision, saliency detection has shown a great success in ranges of visual applications, \eg. object retargeting~\cite{ding2011importance,sun2011scale,Vinyals2014Show}, scene classification~\cite{siagian2007rapid,ren2014region}, visual tracking~\cite{borji2012adaptive,mahadevan2013biologically}, image retrieval~\cite{he2012mobile,gao20123} and semantic segmentation~\cite{donoser2009saliency}.
%In the service of computer vision, recent saliency models have been found to be useful for applications including:
%semantic segmentation, object detection, object proposals, image clustering and retrieval, and cognitive saliency applications such as image captioning and high-level image understanding.
%
Despite decades of valuable research, salient object detection still remains an unsolved research problem because there are large variety of aspects that can contribute to define visual saliency, and it's hard to combine all hand-tuned factors or cues in an appropriate way.

Inspired by human visual attention mechanisms, many early existing methods~\cite{itti98,harel07,Federico12,jiang2013salient,yan2013hierarchical,yang2013saliency} in salient object detection leverage low-level visual features (\eg. color, texture and contrast) with heuristic priors to model and approximate human saliency.
%such as center prior
%
These generic techniques are known to be useful for keeping fine image structures and reducing computation.
Representative methods have set the benchmark on several saliency detection datasets.
%%~\cite{borji2015salient,jiang2013salient}
However, such low-level features and priors can hardly capture high-level semantic knowledge about the object and its surroundings.
Thus, these low-level feature based methods are very far away from distinguishing salient objects from the clutter background and can not generate satisfied predictions. %and understanding the mechanism of human visual attention.
%how saliency works in the context of human perceptions.

In recent years, fully convolutional networks (FCNs), adaptively extracting high-level semantic information from raw images, have shown impressive results in many dense labeling tasks, such as image segmentation~\cite{long2015fully,noh2015learning,dai2016instance}, generic object extraction~\cite{li2016r,hariharan2015hypercolumns}, pose estimation~\cite{yang2016end} and contour detection~\cite{xie2015holistically}.
Motivated by these achievements, several attempts to utilize high-level features of FCNs, have been performed and delivered superior performance in predicting saliency maps~\cite{lee2016deep,li2015visual,liu2016dhsnet,wang2015deep,zhao2015saliency}.
Nevertheless, these state-of-the-art models mainly focus on the non-linear combination of high-level features extracted from the last convolutional layers.
Due to the lack of low-level visual information such as object edge, the predicted results of these methods tend to have poorly localized object boundaries.
%Several approaches have been adopted to handle this problem.
%Therefore, understanding these features and how best to exploit them is of wide applicability.

%succinctly
From above discussions, we note that 1) how to simultaneously utilize multi-level potential saliency cues, 2) how to conveniently find the optimal multi-level feature aggregation strategy, and 3) how to efficiently preserve salient objects' boundaries should become the most intrinsic problems in salient object detection.
To resolve these problems, in this paper, we propose a generic aggregating multi-level convolutional feature framework, namely \textbf{Amulet}, which effectively utilizes multi-level features of FCNs for salient object detection.

Our main contributions are summarized as follows:
\begin{itemize}
\vspace{-2mm}
 \item
We propose a multi-level feature aggregation network, dubbed AmuletNet, which utilizes convolutional features from multiple levels as saliency cues for salient object detection.
AmuletNet integrates multi-level features into multiple resolutions, learns to combine these features at each resolution and predicts saliency maps in a recursive manner.
\vspace{-2mm}
 \item
We propose a deeply recursive supervision learning framework. It effectively incorporates edge-aware feature maps in low-level layers and the predicted results from low resolution features, to achieve accurate object boundary inference and semantic enhancement.
The resulting framework can be trained by end-to-end gradient learning, which uses single-resolution ground truth without additional annotations.
 \vspace{-2mm}
\item
The proposed model (only trained on the MSRA10K dataset~\cite{ChengPAMI}) achieves new state-of-the-art performance on other large-scale salient object detection datasets, including the recent DUTS~\cite{Wang2017CVPR}, DUT-OMRON~\cite{yang2013saliency}, ECSSD~\cite{yan2013hierarchical}, HKU-IS~\cite{zhao2015saliency}, PASCAL-S~\cite{li2014secrets}, SED~\cite{borj2015salient} and SOD~\cite{yan2013hierarchical}.
In addition, the model is fast on modern GPUs, achieving a near real-time speed of 16 fps.
\end{itemize}

%-------------------------------------------------------------------------
\section{Related Work}
%In this section, representative works in salient region detection and  are reviewed. We refer readers to [4] and [5] for a
%survey and a benchmark comparison of the state-of-the-art salient region detection algorithms.
In this section, we briefly review existing representative models for salient object detection.
We also discuss the multi-level feature aggregation methods based on FCNs.
\subsection{Salient object detection}
Over the past decades, lots of salient object detection methods have been developed.
The majority of salient object detection methods are based on low-level hand-crafted features, \eg, image contrast~\cite{Federico12,jiang2013salient}, color~\cite{li2014secrets,borj2015salient}, texture~\cite{yan2013hierarchical,yang2013saliency}.
A complete survey of these methods is beyond the scope of this paper and we refer the readers to a recent survey paper~\cite{borji2015salient} for details.

Recently, deep learning based approaches, in particular the convolutional neural networks (CNNs), have delivered
remarkable performance in many recognition tasks.
%in saliency detection
A lot of research efforts have been made to develop various deep architectures for useful features that characterize salient objects or regions.
For instance, Wang \etal~\cite{wang2015deep} first propose two deep neural networks to integrate local pixel estimation and global proposal search for salient object detection.
%Meanwhile,
Li \etal~\cite{li2015visual} predict the saliency degree of each superpixel by taking multi-scale features in multiple generic CNNs.
Zhao \etal~\cite{zhao2015saliency} also predict the saliency degree of each superpixel by taking global and local context into account, and detect salient objects in a multi-context deep CNN.
Though these methods achieve better results than traditional counterparts, none of them handle low-level details perfectly, and all of their models include several fully connected layers, which are computationally expensive and drop spatial information of input images.
To remedy above problems, Lee \etal~\cite{lee2016deep} propose to encode low-level distance map and high-level sematic features of deep CNNs for salient object detection.
Liu \etal~\cite{liu2016dhsnet} propose a deep hierarchical saliency network to learn enough global structures and progressively refine the details of saliency maps step by step via integrating local context
information.
In addition, Li \etal~\cite{LiYu16} design a pixel-level fully convolutional stream
and a segment-level spatial pooling stream to produce pixel-level saliency predictions.
Wang \etal~\cite{wang2016saliency} develop deep recurrent FCNs to incorporate the coarse predictions as saliency priors and stage-wisely refine the generated predictions.
In contrary to the above methods only used specific-level features, we observe that features from all levels are potential saliency cues and helpful for salient object detection.
%
%Note, however, that replacing it with any other deep network is straightforward.experimentally
In light of this observation, we develop a new multi-level feature aggregation approach based on deep FCNs, and show that beyond refining the predicted saliency map, the approach can also jointly learn to preserve object boundaries.
\subsection{Feature aggregation in FCNs}
%characterize
Several works on visualizing deep CNNs~\cite{simonyan2013deep,zeiler2014visualizing,mahendran2015understanding,Wang2015Visual}  indicate that convolutional features at different levels describe the object and its surroundings from different views.
High-level semantic features helps the category recognition of image regions, while low-level visual features help to generate sharp, detailed boundaries for high-resolution prediction.
%Nevertheless
However, how to effectively and efficiently exploit multi-level convolutional features remains an open question. To this end, several valuble attempts have been performed.
The seminal FCN method~\cite{long2015fully} introduces skip-connections and adds high-level prediction layers to intermediate layers to generate pixel-wise prediction results at multiple resolutions.
The Hypercolumn method~\cite{hariharan2015hypercolumns} also integrates convolutional features from multiple middle layers and learns high-level dense classification layers.
The SegNet~\cite{segnet} and DeconvNet~\cite{noh2015learning} employ a convolutional encoder-decoder network with pooling index guided deconvolution modules to exploit the features from multi-level convolutional layers.
Similarly,~ the U-Net~\cite{ronneberger2015u} apply multiple skip-connections to construct a contracting path to capture context and a symmetric expanding path that enables precise localization.
The HED model~\cite{xie2015holistically} employs deeply supervised structures, and automatically learns rich hierarchical representations that are fused to resolve the challenging ambiguity in edge and object boundary detection.

Our proposed approach clearly differs from the above-mentioned methods in three aspects.
Firstly, our method aggregates multi-level features at multiple resolutions. We use a pre-trained FCN and integrate all level features into multiple resolutions at once. Our method can simultaneously incorporate coarse semantics and fine details. %
Although all above methods seem to be useful for aggregating multi-level features, their aggregation is carried out in a stage-wise manner rather than jointly integrating.
Secondly, our method employs a bidirectional information stream, which facilitates complement effect in prediction.
In contrary, all above-mentioned methods simply aggregate multiple level features from one direction, i.e., low to high or high to low.
Thirdly, our method is able to refine the coarse high-level semantic predictions by exploiting low-level visual features.
In particular, our method employs edge-aware feature maps of low-level layers into the prediction modules which help to preserve objects' boundaries.
\begin{figure*}
\begin{center}
%\fbox{\rule{0pt}{2in} \rule{.9\linewidth}{0pt}}
\includegraphics[width=0.95\linewidth,height=8.0cm]{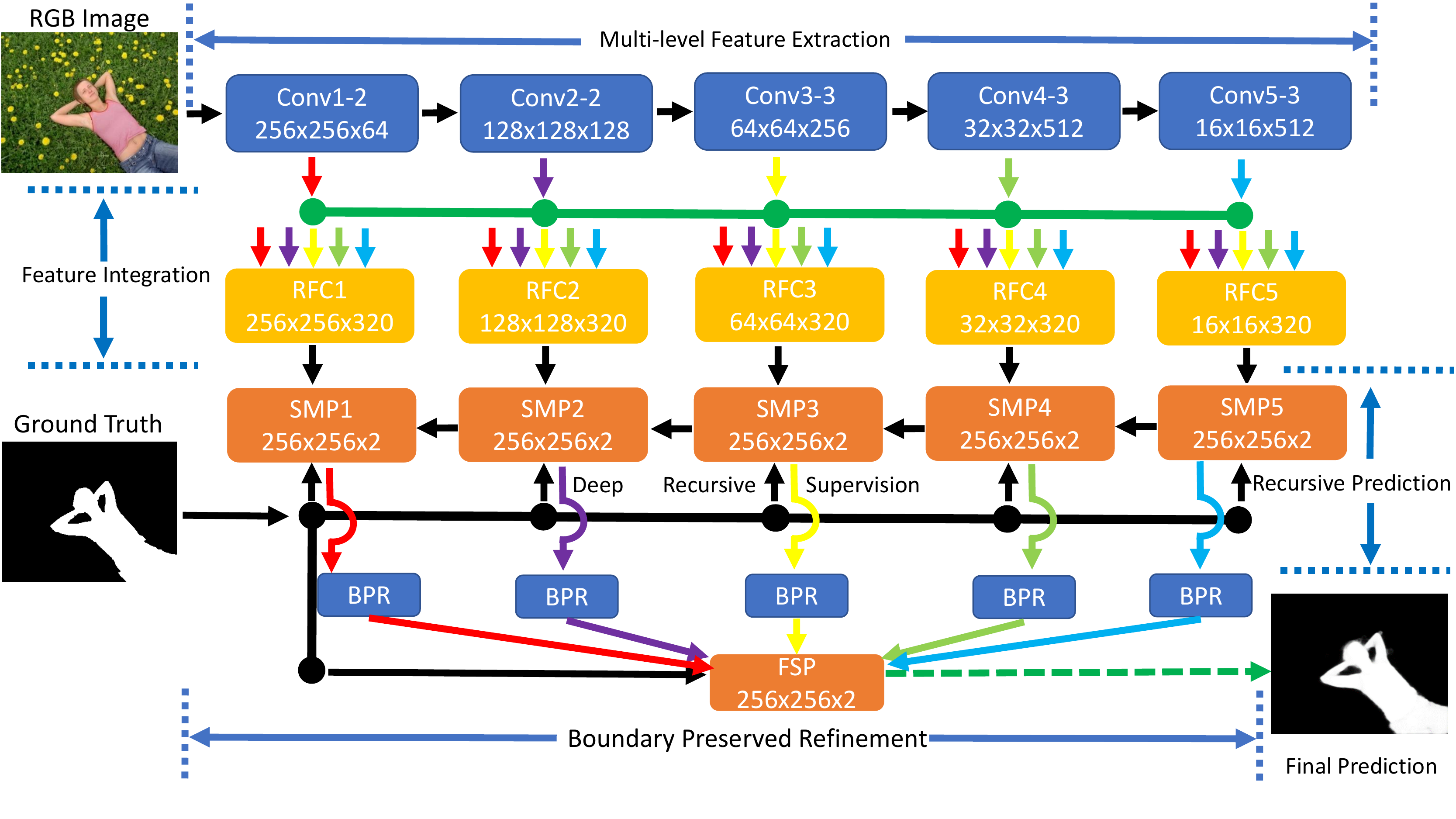}
\end{center}
\vspace{-4mm}
\caption{The overall architecture of our proposed \textbf{Amulet} model. Each colorful box is considered as a feature block. The arrows between blocks indicate the information stream. Given an input image (256$\times$256$\times$3), multi-level features are first generated by the feature extraction network (VGG-16~\cite{simonyan2014very}). Then feature integration is performed by resolution-based feature combination modules (RFCs). After that, deep recursive supervision (DRS) is employed to improve the interaction of multiple predictions. Finally, boundary preserved refinements (BPRs) are used to refine the predicted saliency maps. The final saliency map is the fused output of multiple predicted saliency maps.}
\label{fig:Amulet}
\vspace{-5mm}
\end{figure*}
\section{Aggregating Convolutional Feature Model}
In this section, we begin by describing the components of our proposed AmuletNet architecture in Section 3.1. Then we give the detailed formulas of our bidirectional information aggregating learning method in Section 3.2. In the end, we construct saliency inference based on the multi-level predictions of the proposed \textbf{Amulet}.
\subsection{AmuletNet architecture}
\begin{figure}
\begin{center}
%\fbox{\rule{0pt}{2in} \rule{.9\linewidth}{0pt}}
\includegraphics[width=0.98\linewidth,height=3.5cm]{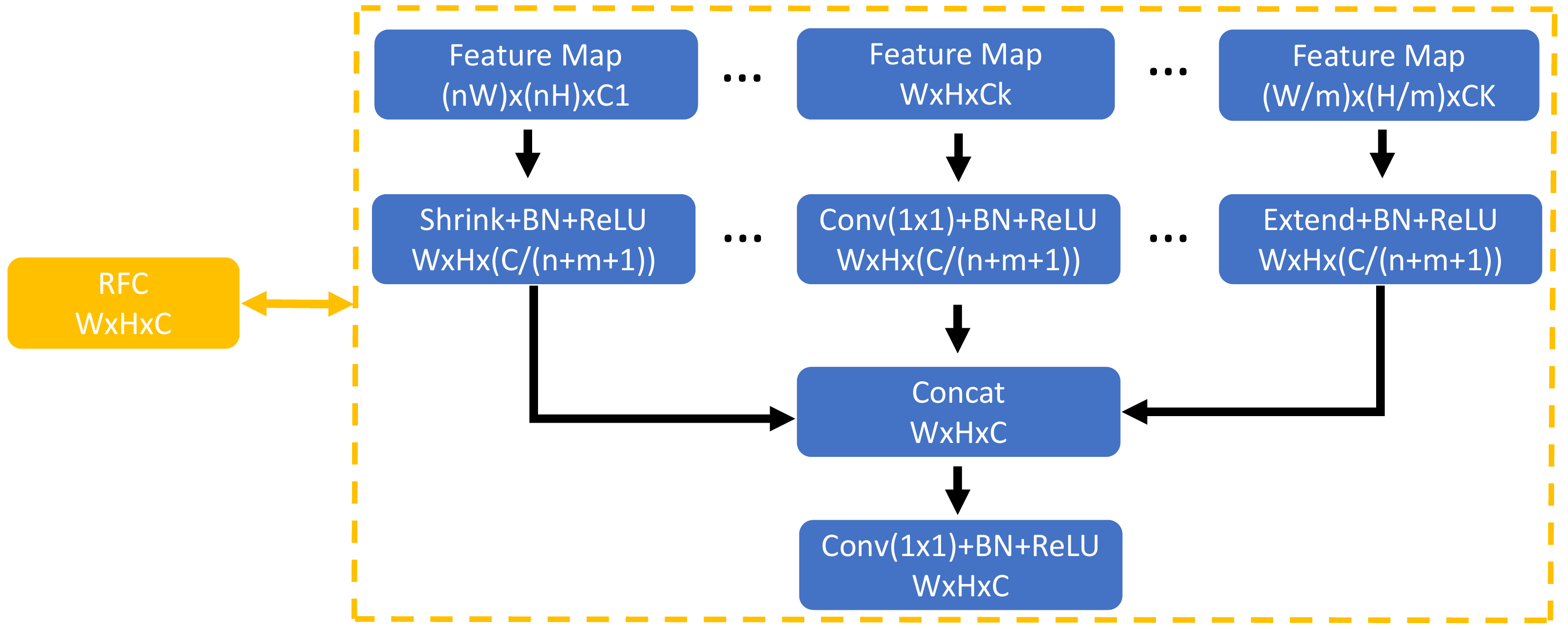}
\end{center}
\vspace{-4mm}
\caption{Details of the RFC module. The RFC first takes feature maps with different resolutions and channels as input. Then shrink and extend operators resize the feature maps to the same spatial resolution and equal channels. Finally, the concatenation and $1\times1$ convolution are used to generate the integrated features.}
\label{fig:RFC}
\vspace{-5mm}
\end{figure}

Our proposed AmuletNet consists of four components: multi-level feature extraction, resolution-based feature integration,~ recursive saliency map prediction~ and~ boundary preserved refinement.
%a feature extraction network that extracts multi-level convolutional features, a feature encoding network that combines multi-level features into multiple resolutions, a recursive prediction network that infers dense prediction from multiple combined features, and boundary preserved connections that boost the localization precision.
%
The four main components are jointly trained to optimize the output saliency detection quality.
The overall architecture is illustrated in Fig.~\ref{fig:Amulet}.
%\vspace{-2mm}
{\flushleft\textbf{Multi-level feature extraction.}}
The first component of our architecture is a deep feature extraction network, which takes the input image and produces feature maps for convolutional feature integration.
We build our architecture on the VGG-16 model from~\cite{simonyan2014very}, which is well known for its elegance and simplicity, and at the same time yields nearly state-of-the-art results in image classification and good generalization properties. In the VGG-16 model there are five max-pooling stages with kernel size 2 and stride 2.
Given an input image with size $W\times H$, the output feature maps have size $\lfloor\frac{W}{2^5},\frac{H}{2^5}\rfloor$, thus a FCN model built upon the VGG-16 would output feature maps reduced by a factor of 32.
%
%Multiple stages of spatial pooling and convolution strides reduce the final image prediction typically by a factor of 32 in each dimension, thereby losing much of the finer image structure.
%
To balance the semantic context and fine image details, we remove the last pooling stage and enlarge the size of the input image.
This way, the output feature maps of our feature extraction network are rescaled by a factor of 16 with respect to the input image.
We take feature maps at five levels from the VGG-16 model: conv1-2 (which contains 64 feature maps), conv2-2 (128 feature maps), conv3-3 (256 feature maps), conv4-3 (512 feature maps) and conv5-3 (512 feature maps).
Note that our feature extraction network is extremely flexible in that it can be replaced and modified in various ways, such as using different layers or networks, \eg. VGG-19~\cite{simonyan2014very} and ResNet~\cite{He2016Deep}.%AlexNet~\cite{Krizhevsky2012ImageNet},
\vspace{-2mm}
{\flushleft \textbf{Resolution-based feature integration.}}
Considering the inconsistent resolution of multi-level convolutional features, we propose a novel resolution-based feature combination structure, named RFC.
The RFC structure consists of both \textbf{shrink} and \textbf{extend} branches.
Assume $\textbf{I}$ is the input image; $\tau=\lfloor\frac{W}{2^l},\frac{H}{2^l}\rfloor$ is the target resolution of integrated feature maps, and identified by feature level $l(=0,1,...,L)$; $\textbf{F}_{n}(\textbf{I})$ denotes a 3D tensor, i.e., the feature maps generated by the feature extraction network with $n\times \tau$ resolution.
Thus, the proposed RFC generates the integrated feature maps by
\begin{equation}
  \label{equ:equ1}
\begin{aligned}
 \textbf{F}^{\tau} = \textbf{W}^{\tau}*\textbf{Cat}(S_{n}(\textbf{F}_{n}(\textbf{I});\psi_{n}),...,S_{1}(\textbf{F}_{1}(\textbf{I});\psi_{1}),\\ E_{1}(\textbf{F}_{1}(\textbf{I});\varphi_{1}),...,E_{m}(\textbf{F}_{m}(\textbf{I});\varphi_{m})),
\end{aligned}
\end{equation}
where $*$ represents convolution operation;$S_{n}(\cdot;\psi_{n})$ denotes the shrink operator parameterized by $\psi_{n}$ that aims to down-sample the input high-resolution feature maps by a factor of $n$, while the extend operator $E_{m}(\cdot;\varphi_{m})$ aims to up-sample the low-resolution ones by a factor of $m$.
The shrink operators can be convolution or pooling. The extend operators can be deconvolution or interpolation.
$\textbf{Cat}$ is the cross-channel concatenation.
$\textbf{W}^{\tau}$ is the parameter for combining the concatenated feature maps.
The details of RFC are shown in Fig.~\ref{fig:RFC}.
For our proposed AmuletNet, we take feature maps at five different levels ($L=4$) from the above feature extraction network.
%
%These feature maps essentially hold different spatial resolutions.
%
We utilize RFCs to resize all level feature maps into the five spatial resolution by performing 64 convolution or deconvolution operations.
The generated features are concatenated into a tensor with 320 channels at each resolution.
Then we use a convolutional layer with $1\times 1$ kernel size to weight the importance of each feature map.
For computational efficiency, 64 convolutional kernels are used to combine each tensor into 64 integrated feature maps.
This way, each integrated feature map will simultaneously incorporate coarse semantics and fine details.
\vspace{-5mm}
{\flushleft \textbf{Recursive saliency map prediction.}}
The integrated feature maps already contains various saliency cues, so we can use them to predict the saliency map.
A direct method is to deconvolute the integrated feature maps at each level into the size of the input image, and add a new convolutional layer to produce the predicted saliency map.
Although this method can detect salient objects from different levels, the inner connection of different-level predictions is missing.
As a result, the independent prediction is not satisfactory enough, both quantitatively and visually, and further
post-processing is needed~\cite{Li2016DeepSaliency,wang2016saliency}.
To facilitate the interaction of multiple predictions, we propose a recursive prediction architecture, i.e. Deep Recursive Supervision (DRS) in Fig.~\ref{fig:Amulet}, to hierarchically and progressively absorb high-level predictions and render pixel-wise supervised information.
The proposed DRS includes saliency map prediction modules (SMP) and the deeply supervised learning mechanism~\cite{xie2015holistically}.
The SMP incorporates autoregressive recurrent connections into the predictions from high-level to low.
In each level $l$, the SMP takes integrated feature maps $\textbf{F}^{\tau}$ and the high-level prediction $\textbf{P}^{l+1}$ as input, and produces the new prediction of this level as
\begin{equation}
  \label{equ:equ2}
\textbf{P}^{l}=
\left\{
\begin{aligned}
&\textbf{W}_{r}*\sigma(\textbf{W}_{F^{\tau}}\star_{s}\textbf{F}^{\tau}+\textbf{W}_{P^{l+1}}*\textbf{P}^{l+1}+\textbf{b}),l<L\\
&\textbf{W}_{F^{\tau}}\star_{s}\textbf{F}^{\tau}+\textbf{b}, l = L
\end{aligned}
\right.
\end{equation}
where $\star_{s}$ represents deconvolution operation with stride $s$ to ensure the same spatial size of the output prediction.
$\textbf{W}_{F^{\tau}}$ and $\textbf{W}_{P^{l+1}}$ are the integrated feature weight and the output prediction weight, respectively.
$\textbf{b}$ is the bias parameter.
$\sigma$ is the ReLU activation function.
$\textbf{W}_{r}$ is the recursive weight.
From Eq.(\ref{equ:equ2}) and Fig.~\ref{fig:Amulet}, we can see that multiple autoregressive recurrent connections
ensure that the new prediction has multiple paths from the input to the output, which facilitates effective information exchanges.
Besides, we employ deeply supervised learning into the SMPs.
This way, the pixel-wise supervised information from ground truth will guide the recursive saliency map prediction at each level, making the SMPs be able to propagate fine details back to the predictions of large contexts.
Thus, DRS can build a bidirectional information stream aggregation, which facilitates complement effect in prediction.
We will fully elaborate the bidirectional information aggregating learning in Section 3.2.
The experiments in Section 4.4 show the superiority of DRS over the deeply supervised learning in~\cite{xie2015holistically}.
\vspace{-5mm}
{\flushleft\textbf{Boundary preserved refinement.}}
To further improve the detection accuracy, we add boundary refinements by introducing short connections to the predicted results.
Our approach bases on the observation that low-level feature maps in the conv1-2 layer have edge-preserving properties~\cite{zeiler2014visualizing,mahendran2015understanding}.
We expect that these low-level features help to predict objects' boundary.
Besides, the features also have the same spatial resolution with respect to the input image.
For boundary refinement, a convolutional layer with $1\times1$ kernel size is first applied to the conv1-2 layer, yielding
boundary predictions $\textbf{B}^{l}$.
Then $\textbf{B}^{l}$ are added to the raw prediction for better aligned object boundaries,
\begin{equation}
  \label{equ:equ3}
\begin{aligned}
 \textbf{P}_{b}^{l} = \textbf{W}_{b}*\sigma(\textbf{B}^{l}+\textbf{P}^{l}),
\end{aligned}
\end{equation}
where $\textbf{W}_{b}$ is the refinement parameter.
ReLU is used so that the boundary prediction is in the range of zero to infinity.
Based on the boundary preserved refinements $\textbf{P}_{b}$, a additional convolutional layer is applied and learned to produce the fusion saliency prediction (FSP) as the final output.
%
%The above method appropriately combine different level predictions such that the most visually distinctive objects can be extracted with fine structures.
\subsection{Bidirectional information aggregating learning}
%\textbf{Training phase}
Given the salient object detection training dataset $ S=\{(X_n,Y_n)\}^{N}_{n=1}$ with $N$ training pairs, where $X_n =
\{x^n_j,j = 1,...,T\}$ and $Y_n = \{y^n_j,j = 1,...,T\}$ are the input image and the binary ground-truth image with $T$ pixels, respectively.
$y^n_j = 1$ denotes the foreground pixel and $y^n_j = 0$ denotes the background pixel.
For notional simplicity, we subsequently drop the subscript $n$ and consider each image independently.
We denote $\textbf{W}$ as the parameters of the feature extraction network and RFCs.
Supposing the network has $M$ predictions, including one fused prediction and $M-1$ specific-level predictions. In our AmuletNet, we have $M =6$.
For the fused prediction, the loss function can be expressed as
\begin{equation}
  \label{equ:equ4}
\begin{aligned}
  \mathcal{L}_f(\textbf{W},w_f)= - \beta \sum_{j\in Y_{+}} \text{log~Pr}(y_{j}=1|X;\textbf{W},w_f)\\
  -(1-\beta)\sum_{j\in Y_{-}} \text{log~Pr}(y_{j}=0|X;\textbf{W},w_f),
\end{aligned}
\end{equation}
where $w_f$ is the classifier parameter for the fused prediction.
$Y_{+}$ and $Y_{-}$ denote the foreground and background label sets, respectively.
The loss weight $\beta = |Y_{+}|/|Y|$, and $|Y_{+}|$ and $|Y_{-}|$ denote the foreground and background pixel number, respectively.
Pr$(y_j =1|X;\textbf{W};w_f)\in [0,1]$ is the confidence score of the fused prediction that measures how likely the pixel belong to the foreground.

For the prediction at level $l$, the loss function can be represented by
\begin{equation}
  \label{equ:equ5}
\begin{aligned}
  \mathcal{L}_l(\textbf{W},\theta_l,w_l)= - \beta \sum_{j\in Y_{+}} \text{log~Pr}(y_{j}=1|X;\textbf{W},\theta_l,w_l)\\
  -(1-\beta)\sum_{j\in Y_{-}} \text{log~Pr}(y_{j}=0|X;\textbf{W},\theta_l,w_l),
\end{aligned}
\end{equation}
where $\theta_l = (w^r_l,w^{b}_l)$ is the parameter of the recursive prediction component and boundary refinement component in the prediction module.
$w_l$ is the classifier parameter for the prediction at level $l$.
Thus, the joint loss function for all predictions is obtained by
\begin{align}
  \mathcal{L}(\textbf{W},\theta,w)= \alpha_f \mathcal{L}_f(\textbf{W},w_f)
  +\sum_{l=0}^{L}\alpha_l\mathcal{L}_l(\textbf{W},\theta_l,w_l),
  \label{equ:equ6}
\end{align}
where $\alpha_f$ and $\alpha_l$ are the loss weights to balance each loss term. For simplicity and fair comparison, we set $\alpha_f=\alpha_l=1$ as used in~\cite{xie2015holistically}.
The above loss function is continuously differentiable, so we can use the stochastic gradient descent (SGD) method to obtain the optimal parameters,
\begin{align}
(\textbf{W}^{*},\theta^{*},w^{*})= \text{arg min}~\mathcal{L}(\textbf{W},\theta,w).
  \label{equ:equ7}
\end{align}

Our aggregating learning method has several significant differences with other deeply supervised implementations, i.e., DHS~\cite{liu2016dhsnet} and HED~\cite{xie2015holistically}.
In DHS and HED, the deep supervision is directly applied on side-outputs, while in our method the deep supervision is applied on multiple same resolution predictions.
According to Eq.(\ref{equ:equ2}), each recursive prediction contains the information of two predictions at least, endowing our method the capability to propagate the supervised information across deep layers in a bidirectional manner.
The bold black arrows in Fig.~\ref{fig:Amulet} illustrate the bidirectional information stream.
Besides, DHS needs to specify scales for side-outputs to minimize the multi-scale error, which requires additional annotation for each scale.
In contrast, the proposed method adaptively unify the scale information into the size of input images, without using multi-scale
annotations.
In addition, different from the methods used sigmoid classifiers in~\cite{liu2016dhsnet,xie2015holistically}, we use the following softmax classifier to evaluate the prediction scores:
\begin{align}
\text{Pr}(y_{j}=1|X;\textbf{W},\theta,w) = \frac{e^{z_1}}{e^{z_0}+e^{z_1}},
  \label{equ:equ8}
\end{align}
\begin{align}
\text{Pr}(y_{j}=0|X;\textbf{W},\theta,w) = \frac{e^{z_0}}{e^{z_0}+e^{z_1}},
  \label{equ:equ9}
\end{align}
where $z_0$ and $z_1$ are the score of each label of training data. In this way, each prediction of the AmultNet is composed of a foreground excitation map ($\textbf{M}^{fe}$) and a background excitation map ($\textbf{M}^{be}$).
We utilize $\textbf{M}^{fe}$ and $\textbf{M}^{be}$ of all-level predictions to generate the final fusion.
This strategy not only increases the pixel-level discrimination but also captures context contrast information.
\subsection{Saliency inference}
Although the architecture we use in this work can produce $M$ predictions computed by Eq.(\ref{equ:equ8}) with the optimal parameters $(\textbf{W}^{*},\theta^{*},w^{*})$, we observe that the quality of the predictions at different levels varies widely.
The more lower level, the better they are.
The fused prediction generally appears much better than other predictions.
For saliency inference, we can simply use the fused prediction as our final saliency map.
However, saliency inference emphasize the contrast between foreground and background.
Therefore, more biologically we utilize the mean contrast of different predictions to further improve the detection accuracy during saliency inference.
%
%We use our proposed excitation map to ca
%
Formally, let $\textbf{M}_{l}^{fe} (\textbf{M}_{f}^{fe})$ and $\textbf{M}_{l}^{be} (\textbf{M}_{f}^{be})$ denote the foreground excitation map and background excitation map at level $l$ (of the fused prediction), respectively.
They can be computed by Eq.(\ref{equ:equ8}) and Eq.(\ref{equ:equ9}).
Thus, the final saliency map can be obtained by
%\begin{align}
%  \textbf{S} & = \textbf{Max}(0,\textbf{Mean}(\sum_{l=0}^{L}(\textbf{M}_{l}^{fe}-\textbf{M}_{l}^{be}))+(\textbf{M}_{f}^{fe}-\textbf{M}_{f}^{be})),
%  \label{equ:equ10}
%\end{align}
\begin{align}
  \textbf{S} & = \sigma(\textbf{Mean}(\sum_{l=0}^{L}(\textbf{M}_{l}^{fe}-\textbf{M}_{l}^{be}))+(\textbf{M}_{f}^{fe}-\textbf{M}_{f}^{be})),
  \label{equ:equ10}
\end{align}
where $\textbf{Mean}$ is the pixel-wise mean and $\sigma$ is the ReLU activation function for clipping the negative values.
%-------------------------------------------------------------------------
\section{Experiments}
%In this section, we first describe the experimental setup.
%
%Then, we evaluate and analyze the proposed EDFCN model on typical saliency detection datasets.
\subsection{Experimental Setup}
\textbf{Datasets:} For the training, we utilize the MSRA10K dataset~\cite{ChengPAMI}, which includes 10,000 images with high quality pixel-wise annotations.
Most of the images in this dataset contain only one salient object.
To improve the varieties, we simply augment this dataset by mirror reflection and rotation techniques ($0^{\circ}, 90^{\circ}, 180^{\circ}, 270^{\circ}$), producing 80,000 training images totally.

For the performance evaluation, we adopt seven public saliency detection datasets as follows.

\textbf{DUT-OMRON}~\cite{yang2013saliency}. This dataset has 5,168 high quality images. Images of this dataset have one or more salient objects and relatively complex background. Thus this dataset is more difficult and challenging, and provides more space of improvement for related research in saliency detection.

\textbf{DUTS}~\cite{zhao2015saliency}. This dataset is currently the largest saliency detection benchmark, and contains 10,553 training images (DUTS-TR) and 5,019 test images (DUTS-TE) with high quality pixel-wise annotations.
%All training images are collected from the ImageNet DET training/val sets [1], while test images are collected from the ImageNet DET test set and the SUN data set [2].
Both the training and test set contain very challenging scenarios for saliency detection.
%To our knowledge, DUTS is currently the largest saliency detection benchmark with the explicit training/test evaluation protocol. For fair comparison in the future research, the training set of DUTS serves as a good candidate for learning DNNs, while the test set and other public data sets can be used for evaluation.

\textbf{ECSSD}~\cite{yan2013hierarchical}. This dataset contains 1,000 natural images, which include many semantically meaningful and complex structures in their ground truth segmentation.

\textbf{HKU-IS}~\cite{zhao2015saliency}. This dataset has 4,447 images with high quality pixel-wise annotations.
Images of this dataset are well chosen to include multiple disconnected salient objects or objects touching the image boundary.

\textbf{PASCAL-S}~\cite{li2014secrets}. This dataset is generated from the PASCAL VOC dataset~\cite{Everingham2010ThePV} and contains 850 natural images.

\textbf{SED}~\cite{borj2015salient}. This dataset contains two subsets: \textbf{SED1} and \textbf{SED2}.
The \textbf{SED1} has 100 images each containing only one salient object, while the \textbf{SED2} has 100 images each containing two salient objects.

\textbf{SOD}~\cite{yan2013hierarchical}. This dataset has 300 images, and it was originally designed for image segmentation.
Pixel-wise annotation of salient objects was generated by~\cite{jiang2013salient}.
This dataset is challenging since many images contain multiple objects either with low contrast or touching the image boundary.

\textbf{Implementation Details:}
We implement our approach based on the MATLAB R2014b platform with the Caffe toolbox~\cite{jia2014caffe}.
We run our approach in a quad-core PC machine with an i7-4790 CPU (with 16G memory) and a NVIDIA Titan X GPU (with 12G memory).
We train our model using augmented images from the MSRA10K dataset. We do not use validation set and train the model
until its training loss converges.
The parameters of multi-level feature extraction layers are initialized from the VGG-16 model~\cite{simonyan2014very}.
For other convolutional layers, we initialize the weights by the ``msra'' method~\cite{He2015Delving}.
We use the SGD method to train our network with a momentum 0.9 and a weight decay 0.0001.
We set the base learning rate to 1e-8 and decrease the learning rate by 10\% when training loss reaches a flat.
The training process takes almost 16 hours and converges after 200k iterations with mini-batch size 8.
When testing, the proposed salient object detection algorithm runs at about \textbf{16 fps} with $256\times256$ resolution.
The source code can be found at \textcolor[rgb]{1,0,0}{http://ice.dlut.edu.cn/lu/}.

\textbf{Evaluation Metrics:}
We utilize three main metrics to evaluate the performance of different salient object detection algorithms, including the precision-recall (PR) curves, F-measure and mean absolute error (MAE)~\cite{borji2015salient}.
The precision and recall are computed by thresholding the predicted saliency map, and comparing the binary map with the ground truth.
The PR curve of a dataset demonstrates the mean precision and recall of saliency maps at different thresholds.
The F-measure is a harmonic mean of average precision and average recall, and can be calculated by
\vspace{-0.5mm}
\begin{align}
  F_{\beta} =\frac{(1+\beta^2)\times Precision\times Recall}{\beta^2\times Precision \times Recall}.
    \label{equ:equ19}
\end{align}
\vspace{-0.5mm}
We set $\beta^2$ to be 0.3 to weigh precision more than recall as suggested in~\cite{yan2013hierarchical}~\cite{wang2015deep}~\cite{borji2015salient}~\cite{yang2013saliency}.

We report the performance when each saliency map is binarized with an image-dependent threshold.
The threshold is determined to be twice the mean saliency of the image:
\begin{align}
T = \frac{2}{W\times H}\sum_{x=1}^{W}\sum_{y=1}^{H}S(x,y),
  \label{equ:equ3}
\end{align}
where $W$ and $H$ are width and height of an image, $S(x,y)$ is the saliency value of the pixel at
$(x,y)$. We report the average precision, recall and F-measure over each dataset.

The above overlapping-based evaluations usually give higher score to methods which assign high saliency score to salient pixel correctly.
However, the evaluation on non-salient regions can be unfair especially for the methods which successfully detect non-salient regions, but miss the detection of salient regions.
Therefore, we also calculate the mean absolute error (MAE) for fair comparisons as suggested by~\cite{borji2015salient}.
The MAE evaluates the saliency detection accuracy by
\vspace{-1mm}
\begin{align}
MAE = \frac{1}{W\times H}\sum_{x=1}^{W}\sum_{y=1}^{H}|S(x,y)-G(x,y)|,
  \label{equ:equ3}
\end{align}
\vspace{-0.5mm}
where $G$ is the binary ground truth mask.
%---------------------------------------------------------------
\subsection{Performance Comparison with State-of-the-art}
%, DHS~\cite{liu2016dhsnet}
We compare our algorithm with other 11 state-of-the-art ones including 7 deep learning based algorithms (DCL~\cite{LiYu16}, DHS~\cite{liu2016dhsnet}, DS~\cite{Li2016DeepSaliency}, ELD~\cite{lee2016deep}, LEGS~\cite{wang2015deep}, MDF~\cite{zhao2015saliency}, RFCN~\cite{wang2016saliency})
and 4 conventional algorithms (BL\cite{tong2015bootstrap}, BSCA~\cite{qin2015saliency}, DRFI~\cite{jiang2013salient}, DSR~\cite{li2013saliency}).
For fair comparison, we use either the implementations with recommended parameter settings or the saliency maps provided by the authors.
%%%%%%%%%%%%%%%%%%%%%%%%%%%%%%%%%%%%%%%%%%%%%%------------------------------------------------
\setlength{\tabcolsep}{2.95pt}
\begin{table*}
\vspace{-5mm}
\begin{center}
\doublerulesep=0.6pt
\begin{tabular}{|c|c|c|c|c|c|c|c|c|c|c|c|c|c|c|c|c|c|c|c|c|c|c|c|c|||c|c|c|c|c|c|c|c|||}
\hline
\multicolumn{4}{|c|}{}
&\multicolumn{4}{|c|}{DUT-OMRON}
&\multicolumn{4}{|c|}{DUTS-TE}
&\multicolumn{4}{|c|}{ECSSD}
&\multicolumn{4}{|c|}{HKU-IS}
&\multicolumn{4}{|c|}{PASCAL-S}
%&\multicolumn{4}{|c|}{SED1}
%&\multicolumn{4}{|c|}{SED2}
&\multicolumn{4}{|c|}{SOD}
\\
\hline
\multicolumn{4}{|c|}{Methods}
&\multicolumn{2}{|c|}{$F_\beta$}&\multicolumn{2}{|c|}{$MAE$}%DUT-OMRON
&\multicolumn{2}{|c|}{$F_\beta$}&\multicolumn{2}{|c|}{$MAE$}%DUTS-TE
&\multicolumn{2}{|c|}{$F_\beta$}&\multicolumn{2}{|c|}{$MAE$}%ECSSD
&\multicolumn{2}{|c|}{$F_\beta$}&\multicolumn{2}{|c|}{$MAE$}%HKU-IS
&\multicolumn{2}{|c|}{$F_\beta$}&\multicolumn{2}{|c|}{$MAE$}%PASCAL-S
%&\multicolumn{2}{|c|}{$F_\beta$}&\multicolumn{2}{|c|}{$MAE$}%SED1
%&\multicolumn{2}{|c|}{$F_\beta$}&\multicolumn{2}{|c|}{$MAE$}%SED2
&\multicolumn{2}{|c|}{$F_\beta$}&\multicolumn{2}{|c|}{$MAE$}%SOD
%%%%%%%%%%%%%%%%%%%%%%%%%%%%% 288x288 %%%%%%%%%%%%%%%%%%%%%%%%%%%
\\
\hline
\multicolumn{4}{|c|}{\textbf{Amulet}}
&\multicolumn{2}{|c|}{\textcolor[rgb]{0,1,0}{0.6471}}&\multicolumn{2}{|c|}{\textcolor[rgb]{0,0,1}{0.09761}}%DUT-OMRON
&\multicolumn{2}{|c|}{\textcolor[rgb]{1,0,0}{0.7365}}&\multicolumn{2}{|c|}{\textcolor[rgb]{0,1,0}{0.08517}}%DUTS-TE
&\multicolumn{2}{|c|}{\textcolor[rgb]{1,0,0}{0.8684}}&\multicolumn{2}{|c|}{\textcolor[rgb]{1,0,0}{0.05874}}%ECSSD
&\multicolumn{2}{|c|}{\textcolor[rgb]{1,0,0}{0.8542}}&\multicolumn{2}{|c|}{\textcolor[rgb]{1,0,0}{0.05214}}%HKU-IS
&\multicolumn{2}{|c|}{\textcolor[rgb]{0,0,1}{0.7632}}&\multicolumn{2}{|c|}{\textcolor[rgb]{0,1,0}{0.09824}}%PASCAL-S
%&\multicolumn{2}{|c|}{\textcolor[rgb]{0,1,0}{0.8917}}&\multicolumn{2}{|c|}{\textcolor[rgb]{1,0,0}{0.06019}}%SED1
%&\multicolumn{2}{|c|}{0.8298}&\multicolumn{2}{|c|}{0.06204}%SED2
&\multicolumn{2}{|c|}{\textcolor[rgb]{0,1,0}{0.7547}}&\multicolumn{2}{|c|}{\textcolor[rgb]{0,1,0}{0.13998}}%SOD
\\
\multicolumn{4}{|c|}{\textbf{Amulet}-1/1}
&\multicolumn{2}{|c|}{0.6413}&\multicolumn{2}{|c|}{0.10161}%DUT-OMRON
&\multicolumn{2}{|c|}{\textcolor[rgb]{0,0,1}{0.7320}}&\multicolumn{2}{|c|}{\textcolor[rgb]{0,0,1}{0.08796}}%DUTS-TE
&\multicolumn{2}{|c|}{\textcolor[rgb]{0,1,0}{0.8678}}&\multicolumn{2}{|c|}{\textcolor[rgb]{0,0,1}{0.05997}}%ECSSD
&\multicolumn{2}{|c|}{\textcolor[rgb]{0,0,1}{0.8460}}&\multicolumn{2}{|c|}{\textcolor[rgb]{0,0,1}{0.05416}}%HKU-IS
&\multicolumn{2}{|c|}{\textcolor[rgb]{0,1,0}{0.7634}}&\multicolumn{2}{|c|}{\textcolor[rgb]{0,0,1}{0.09948}}%PASCAL-S
%&\multicolumn{2}{|c|}{\textcolor[rgb]{1,0,0}{0.8918}}&\multicolumn{2}{|c|}{\textcolor[rgb]{0,1,0}{0.06087}}%SED1
%&\multicolumn{2}{|c|}{\textcolor[rgb]{1,0,0}{0.8333}}&\multicolumn{2}{|c|}{\textcolor[rgb]{1,0,0}{0.06104}}%SED2
&\multicolumn{2}{|c|}{\textcolor[rgb]{0,0,1}{0.7512}}&\multicolumn{2}{|c|}{\textcolor[rgb]{0,0,1}{0.14169}}%SOD
\\
\multicolumn{4}{|c|}{\textbf{Amulet}-1/2}
&\multicolumn{2}{|c|}{0.6408}&\multicolumn{2}{|c|}{0.10178}%DUT-OMRON
&\multicolumn{2}{|c|}{0.7210}&\multicolumn{2}{|c|}{0.08807}%DUTS-TE
&\multicolumn{2}{|c|}{\textcolor[rgb]{0,0,1}{0.8675}}&\multicolumn{2}{|c|}{0.05998}%ECSSD
&\multicolumn{2}{|c|}{0.8456}&\multicolumn{2}{|c|}{0.05421}%HKU-IS
&\multicolumn{2}{|c|}{0.7629}&\multicolumn{2}{|c|}{0.09965}%PASCAL-S
%&\multicolumn{2}{|c|}{\textcolor[rgb]{0,0,1}{0.8914}}&\multicolumn{2}{|c|}{\textcolor[rgb]{0,0,1}{0.06095}}%SED1
%&\multicolumn{2}{|c|}{\textcolor[rgb]{0,1,0}{0.8327}}&\multicolumn{2}{|c|}{\textcolor[rgb]{0,0,1}{0.06112}}%SED2
&\multicolumn{2}{|c|}{0.7509}&\multicolumn{2}{|c|}{0.14177}%SOD
\\
\multicolumn{4}{|c|}{\textbf{Amulet}-1/4}
&\multicolumn{2}{|c|}{0.6392}&\multicolumn{2}{|c|}{0.10219}%DUT-OMRON
&\multicolumn{2}{|c|}{0.7169}&\multicolumn{2}{|c|}{0.08851}%DUTS-TE
&\multicolumn{2}{|c|}{0.8659}&\multicolumn{2}{|c|}{0.06039}%ECSSD
&\multicolumn{2}{|c|}{0.8439}&\multicolumn{2}{|c|}{0.05465}%HKU-IS
&\multicolumn{2}{|c|}{0.7615}&\multicolumn{2}{|c|}{0.10001}%PASCAL-S
%&\multicolumn{2}{|c|}{0.8904}&\multicolumn{2}{|c|}{0.06124}%SED1
%&\multicolumn{2}{|c|}{\textcolor[rgb]{0,0,1}{0.8312}}&\multicolumn{2}{|c|}{\textcolor[rgb]{0,1,0}{0.06142}}%SED2
&\multicolumn{2}{|c|}{0.7503}&\multicolumn{2}{|c|}{0.14204}%SOD
\\
\multicolumn{4}{|c|}{\textbf{Amulet}-1/8}
&\multicolumn{2}{|c|}{0.6356}&\multicolumn{2}{|c|}{0.10282}%DUT-OMRON
&\multicolumn{2}{|c|}{0.6942}&\multicolumn{2}{|c|}{0.08933}%DUTS-TE
&\multicolumn{2}{|c|}{0.8625}&\multicolumn{2}{|c|}{0.06137}%ECSSD
&\multicolumn{2}{|c|}{0.8397}&\multicolumn{2}{|c|}{0.05570}%HKU-IS
&\multicolumn{2}{|c|}{0.7584}&\multicolumn{2}{|c|}{0.10067}%PASCAL-S
%&\multicolumn{2}{|c|}{0.8871}&\multicolumn{2}{|c|}{0.06212}%SED1
%&\multicolumn{2}{|c|}{0.8243}&\multicolumn{2}{|c|}{0.06265}%SED2
&\multicolumn{2}{|c|}{0.7492}&\multicolumn{2}{|c|}{0.14262}%SOD
\\
\multicolumn{4}{|c|}{\textbf{Amulet}-1/16}
&\multicolumn{2}{|c|}{0.6266}&\multicolumn{2}{|c|}{0.10280}%DUT-OMRON
&\multicolumn{2}{|c|}{0.6891}&\multicolumn{2}{|c|}{0.09110}%DUTS-TE
&\multicolumn{2}{|c|}{0.8523}&\multicolumn{2}{|c|}{0.06477}%ECSSD
&\multicolumn{2}{|c|}{0.8327}&\multicolumn{2}{|c|}{0.05821}%HKU-IS
&\multicolumn{2}{|c|}{0.7469}&\multicolumn{2}{|c|}{0.10273}%PASCAL-S
%&\multicolumn{2}{|c|}{0.8805}&\multicolumn{2}{|c|}{0.06424}%SED1
%&\multicolumn{2}{|c|}{0.8109}&\multicolumn{2}{|c|}{0.06695}%SED2
&\multicolumn{2}{|c|}{0.7421}&\multicolumn{2}{|c|}{0.14495}%SOD
\\
\multicolumn{4}{|c|}{\textbf{Amulet}$_{BPR^{-}}$}
&\multicolumn{2}{|c|}{0.6301}&\multicolumn{2}{|c|}{0.12062}%DUT-OMRON
&\multicolumn{2}{|c|}{0.6912}&\multicolumn{2}{|c|}{0.09761}%DUTS-TE
&\multicolumn{2}{|c|}{0.8647}&\multicolumn{2}{|c|}{0.06572}%ECSSD
&\multicolumn{2}{|c|}{0.8402}&\multicolumn{2}{|c|}{0.06302}%HKU-IS
&\multicolumn{2}{|c|}{0.7533}&\multicolumn{2}{|c|}{0.1240}%PASCAL-S
%&\multicolumn{2}{|c|}{0.8725}&\multicolumn{2}{|c|}{0.06610}%SED1
%&\multicolumn{2}{|c|}{0.8143}&\multicolumn{2}{|c|}{0.08651}%SED2
&\multicolumn{2}{|c|}{0.7201}&\multicolumn{2}{|c|}{0.15340}%SOD
\\
\hline
\multicolumn{4}{|c|}{\textbf{DCL}~\cite{LiYu16}}
&\multicolumn{2}{|c|}{\textcolor[rgb]{1,0,0}{0.6842}}&\multicolumn{2}{|c|}{0.15726}%DUT-OMRON
&\multicolumn{2}{|c|}{0.7141}&\multicolumn{2}{|c|}{0.14928}%DUTS-TE
&\multicolumn{2}{|c|}{0.8293}&\multicolumn{2}{|c|}{0.14949}%ECSSD
&\multicolumn{2}{|c|}{0.8533}&\multicolumn{2}{|c|}{0.13587}%HKU-IS
&\multicolumn{2}{|c|}{0.7141}&\multicolumn{2}{|c|}{0.18073}%PASCAL-S
%&\multicolumn{2}{|c|}{0.8546}&\multicolumn{2}{|c|}{0.15131}%SED1
%&\multicolumn{2}{|c|}{0.7946}&\multicolumn{2}{|c|}{0.15652}%SED2
&\multicolumn{2}{|c|}{0.7413}&\multicolumn{2}{|c|}{0.19383}%SOD
\\
\multicolumn{4}{|c|}{\textbf{DHS}~\cite{liu2016dhsnet}}
&\multicolumn{2}{|c|}{-}&\multicolumn{2}{|c|}{-}%DUT-OMRON
%&\multicolumn{2}{|c|}{-}&\multicolumn{2}{|c|}{0.02734}%DUT-OMRON
&\multicolumn{2}{|c|}{\textcolor[rgb]{0,1,0}{0.7301}}&\multicolumn{2}{|c|}{\textcolor[rgb]{1,0,0}{0.06578}}%DUTS-TE
&\multicolumn{2}{|c|}{0.8675}&\multicolumn{2}{|c|}{\textcolor[rgb]{0,1,0}{0.05948}}%ECSSD
&\multicolumn{2}{|c|}{\textcolor[rgb]{0,1,0}{0.8541}}&\multicolumn{2}{|c|}{\textcolor[rgb]{0,1,0}{0.05308}}%HKU-IS
&\multicolumn{2}{|c|}{\textcolor[rgb]{1,0,0}{0.7741}}&\multicolumn{2}{|c|}{\textcolor[rgb]{1,0,0}{0.09426}}%PASCAL-S
%&\multicolumn{2}{|c|}{0.8773}&\multicolumn{2}{|c|}{0.05404}%SED1
%&\multicolumn{2}{|c|}{0.8239}&\multicolumn{2}{|c|}{0.07886}%SED2
&\multicolumn{2}{|c|}{\textcolor[rgb]{1,0,0}{0.7746}}&\multicolumn{2}{|c|}{\textcolor[rgb]{1,0,0}{0.12840}}%SOD
%\\
%\multicolumn{4}{|c|}{\textbf{DHS}~\cite{liu2016dhsnet}}
%%&\multicolumn{2}{|c|}{0.6842}&\multicolumn{2}{|c|}{0.028223}%DUT-OMRON use 3500 images as training data
%&\multicolumn{2}{|c|}{0.8723}&\multicolumn{2}{|c|}{0.060143}%ECSSD
%&\multicolumn{2}{|c|}{\textcolor[rgb]{1,0,0}{0.8533}}&\multicolumn{2}{|c|}{\textcolor[rgb]{0,0,1}{0.054015}}%HKU-IS
%&\multicolumn{2}{|c|}{0.7141}&\multicolumn{2}{|c|}{0.095004}%PASCAL-S
%&\multicolumn{2}{|c|}{0.8546}&\multicolumn{2}{|c|}{0.055249}%SED1
%&\multicolumn{2}{|c|}{0.7946}&\multicolumn{2}{|c|}{0.079836}%SED2
%%&\multicolumn{2}{|c|}{0.7413}&\multicolumn{2}{|c|}{0.128076}%SOD
\\
\multicolumn{4}{|c|}{\textbf{DS}~\cite{Li2016DeepSaliency}}
&\multicolumn{2}{|c|}{0.6028}&\multicolumn{2}{|c|}{0.12038}%DUT-OMRON
&\multicolumn{2}{|c|}{0.6323}&\multicolumn{2}{|c|}{0.09070}%DUTS-TE
&\multicolumn{2}{|c|}{0.8255}&\multicolumn{2}{|c|}{0.12157}%ECSSD
&\multicolumn{2}{|c|}{0.7851}&\multicolumn{2}{|c|}{0.07797}%HKU-IS
&\multicolumn{2}{|c|}{0.6590}&\multicolumn{2}{|c|}{0.17597}%PASCAL-S
%&\multicolumn{2}{|c|}{0.8445}&\multicolumn{2}{|c|}{0.09306}%SED1
%&\multicolumn{2}{|c|}{0.7541}&\multicolumn{2}{|c|}{0.12330}%SED2
&\multicolumn{2}{|c|}{0.6981}&\multicolumn{2}{|c|}{0.18894}%SOD
\\
\multicolumn{4}{|c|}{\textbf{ELD}~\cite{lee2016deep}}
&\multicolumn{2}{|c|}{0.6109}&\multicolumn{2}{|c|}{\textcolor[rgb]{0,1,0}{0.09240}}%DUT-OMRON
&\multicolumn{2}{|c|}{0.6277}&\multicolumn{2}{|c|}{0.09761}%DUTS-TE
&\multicolumn{2}{|c|}{0.8102}&\multicolumn{2}{|c|}{0.07955}%ECSSD
&\multicolumn{2}{|c|}{0.7694}&\multicolumn{2}{|c|}{0.07414}%HKU-IS
&\multicolumn{2}{|c|}{0.7180}&\multicolumn{2}{|c|}{0.12324}%PASCAL-S
%&\multicolumn{2}{|c|}{0.8715}&\multicolumn{2}{|c|}{0.06704}%SED1
%&\multicolumn{2}{|c|}{0.7591}&\multicolumn{2}{|c|}{0.10282}%SED2
&\multicolumn{2}{|c|}{0.7116}&\multicolumn{2}{|c|}{0.15452}%SOD
\\
\multicolumn{4}{|c|}{\textbf{LEGS}~\cite{wang2015deep}}
&\multicolumn{2}{|c|}{0.5915}&\multicolumn{2}{|c|}{0.13335}%DUT-OMRON
&\multicolumn{2}{|c|}{0.5846}&\multicolumn{2}{|c|}{0.13793}%DUTS-TE
&\multicolumn{2}{|c|}{0.7853}&\multicolumn{2}{|c|}{0.11799}%ECSSD
&\multicolumn{2}{|c|}{0.7228}&\multicolumn{2}{|c|}{0.11934}%HKU-IS
&\multicolumn{2}{|c|}{-}&\multicolumn{2}{|c|}{-}%PASCAL-S
%&\multicolumn{2}{|c|}{0.8542}&\multicolumn{2}{|c|}{0.10340}%SED1
%&\multicolumn{2}{|c|}{0.7358}&\multicolumn{2}{|c|}{0.12356}%SED2
&\multicolumn{2}{|c|}{0.6834}&\multicolumn{2}{|c|}{0.19548}%SOD
\\
\multicolumn{4}{|c|}{\textbf{MDF}~\cite{zhao2015saliency}}
&\multicolumn{2}{|c|}{\textcolor[rgb]{0,0,1}{0.6442}}&\multicolumn{2}{|c|}{\textcolor[rgb]{1,0,0}{0.09156}}%DUT-OMRON
&\multicolumn{2}{|c|}{0.6732}&\multicolumn{2}{|c|}{0.09986}%DUTS-TE
&\multicolumn{2}{|c|}{0.8070}&\multicolumn{2}{|c|}{0.10491}%ECSSD
&\multicolumn{2}{|c|}{0.8006}&\multicolumn{2}{|c|}{0.09573}%HKU-IS
&\multicolumn{2}{|c|}{0.7087}&\multicolumn{2}{|c|}{0.14579}%PASCAL-S
%&\multicolumn{2}{|c|}{0.8419}&\multicolumn{2}{|c|}{0.09893}%SED1
%&\multicolumn{2}{|c|}{0.8003}&\multicolumn{2}{|c|}{0.10136}%SED2
&\multicolumn{2}{|c|}{0.7205}&\multicolumn{2}{|c|}{0.16394}%SOD
\\
\multicolumn{4}{|c|}{\textbf{RFCN}~\cite{wang2016saliency}}
&\multicolumn{2}{|c|}{0.6265}&\multicolumn{2}{|c|}{0.11051}%DUT-OMRON
&\multicolumn{2}{|c|}{0.7120}&\multicolumn{2}{|c|}{0.09003}%DUTS-TE
&\multicolumn{2}{|c|}{0.8340}&\multicolumn{2}{|c|}{0.10690}%ECSSD
&\multicolumn{2}{|c|}{0.8349}&\multicolumn{2}{|c|}{0.08891}%HKU-IS
&\multicolumn{2}{|c|}{0.7512}&\multicolumn{2}{|c|}{0.13241}%PASCAL-S
%&\multicolumn{2}{|c|}{0.8502}&\multicolumn{2}{|c|}{0.11662}%SED1
%&\multicolumn{2}{|c|}{0.7667}&\multicolumn{2}{|c|}{0.11306}%SED2
&\multicolumn{2}{|c|}{0.7426}&\multicolumn{2}{|c|}{0.16919}%SOD
\\
\hline
\multicolumn{4}{|c|}{\textbf{BL}~\cite{tong2015bootstrap}}
&\multicolumn{2}{|c|}{0.4988}&\multicolumn{2}{|c|}{0.23881}%DUT-OMRON
&\multicolumn{2}{|c|}{0.4897}&\multicolumn{2}{|c|}{0.23794}%DUTS-TE
&\multicolumn{2}{|c|}{0.6841}&\multicolumn{2}{|c|}{0.21591}%ECSSD
&\multicolumn{2}{|c|}{0.6597}&\multicolumn{2}{|c|}{0.20708}%HKU-IS
&\multicolumn{2}{|c|}{0.5742}&\multicolumn{2}{|c|}{0.24871}%PASCAL-S
%&\multicolumn{2}{|c|}{0.7675}&\multicolumn{2}{|c|}{0.18489}%SED1
%&\multicolumn{2}{|c|}{0.7047}&\multicolumn{2}{|c|}{0.18564}%SED2
&\multicolumn{2}{|c|}{0.5798}&\multicolumn{2}{|c|}{0.26681}%SOD
\\
\multicolumn{4}{|c|}{\textbf{BSCA}~\cite{qin2015saliency}}
&\multicolumn{2}{|c|}{0.5091}&\multicolumn{2}{|c|}{0.19024}%DUT-OMRON
&\multicolumn{2}{|c|}{0.4996}&\multicolumn{2}{|c|}{0.19614}%DUTS-TE
&\multicolumn{2}{|c|}{0.7048}&\multicolumn{2}{|c|}{0.18211}%ECSSD
&\multicolumn{2}{|c|}{0.6544}&\multicolumn{2}{|c|}{0.17480}%HKU-IS
&\multicolumn{2}{|c|}{0.6006}&\multicolumn{2}{|c|}{0.22286}%PASCAL-S
%&\multicolumn{2}{|c|}{0.8048}&\multicolumn{2}{|c|}{0.15347}%SED1
%&\multicolumn{2}{|c|}{0.7062}&\multicolumn{2}{|c|}{0.15784}%SED2
&\multicolumn{2}{|c|}{0.5835}&\multicolumn{2}{|c|}{0.25135}%SOD
\\
\multicolumn{4}{|c|}{\textbf{DRFI}~\cite{jiang2013salient}}
&\multicolumn{2}{|c|}{0.5504}&\multicolumn{2}{|c|}{0.13777}%DUT-OMRON
&\multicolumn{2}{|c|}{0.5407}&\multicolumn{2}{|c|}{0.17461}%DUTS-TE
&\multicolumn{2}{|c|}{0.7331}&\multicolumn{2}{|c|}{0.16422}%ECSSD
&\multicolumn{2}{|c|}{0.7218}&\multicolumn{2}{|c|}{0.14453}%HKU-IS
&\multicolumn{2}{|c|}{0.6182}&\multicolumn{2}{|c|}{0.20651}%PASCAL-S
%&\multicolumn{2}{|c|}{0.8068}&\multicolumn{2}{|c|}{0.14799}%SED1
%&\multicolumn{2}{|c|}{0.7341}&\multicolumn{2}{|c|}{0.13336}%SED2
&\multicolumn{2}{|c|}{0.6343}&\multicolumn{2}{|c|}{0.22377}%SOD
\\
\multicolumn{4}{|c|}{\textbf{DSR}~\cite{li2013saliency}}
&\multicolumn{2}{|c|}{0.5242}&\multicolumn{2}{|c|}{0.13886}%DUT-OMRON
&\multicolumn{2}{|c|}{0.5182}&\multicolumn{2}{|c|}{0.14548}%DUTS-TE
&\multicolumn{2}{|c|}{0.6621}&\multicolumn{2}{|c|}{0.17837}%ECSSD
&\multicolumn{2}{|c|}{0.6772}&\multicolumn{2}{|c|}{0.14219}%HKU-IS
&\multicolumn{2}{|c|}{0.5575}&\multicolumn{2}{|c|}{0.21488}%PASCAL-S
%&\multicolumn{2}{|c|}{0.7909}&\multicolumn{2}{|c|}{0.15794}%SED1
%&\multicolumn{2}{|c|}{0.7116}&\multicolumn{2}{|c|}{0.14063}%SED2
&\multicolumn{2}{|c|}{0.5962}&\multicolumn{2}{|c|}{0.23394}%SOD
\\
\hline
\end{tabular}
\vspace{1mm}
\caption{The F-measure and MAE of different saliency detection methods on six large-scale saliency detection datasets. The best three results are shown in \textcolor[rgb]{1,0,0}{red},~\textcolor[rgb]{0,1,0}{green} and \textcolor[rgb]{0,0,1}{blue}. The proposed methods rank first or second on these datasets.}
\vspace{-7mm}
\label{table:fauc}
\end{center}
\end{table*}
%-------------------------------------------------------------------------
\begin{figure*}
\begin{center}
\begin{tabular}{@{}c@{}c@{}c@{}c}
\includegraphics[width=0.245\linewidth,height=3.85cm]{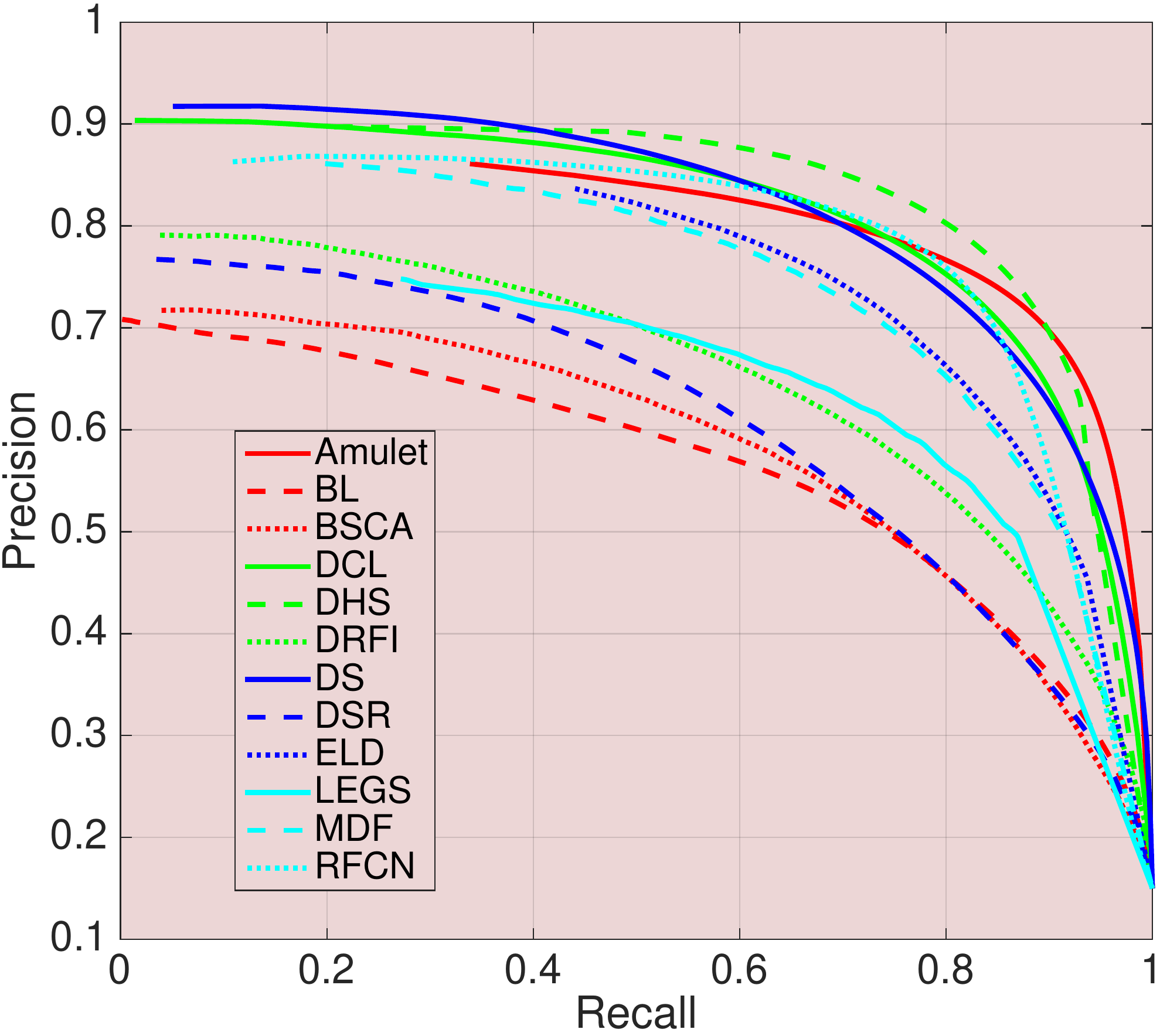} \ &
\includegraphics[width=0.245\linewidth,height=3.85cm]{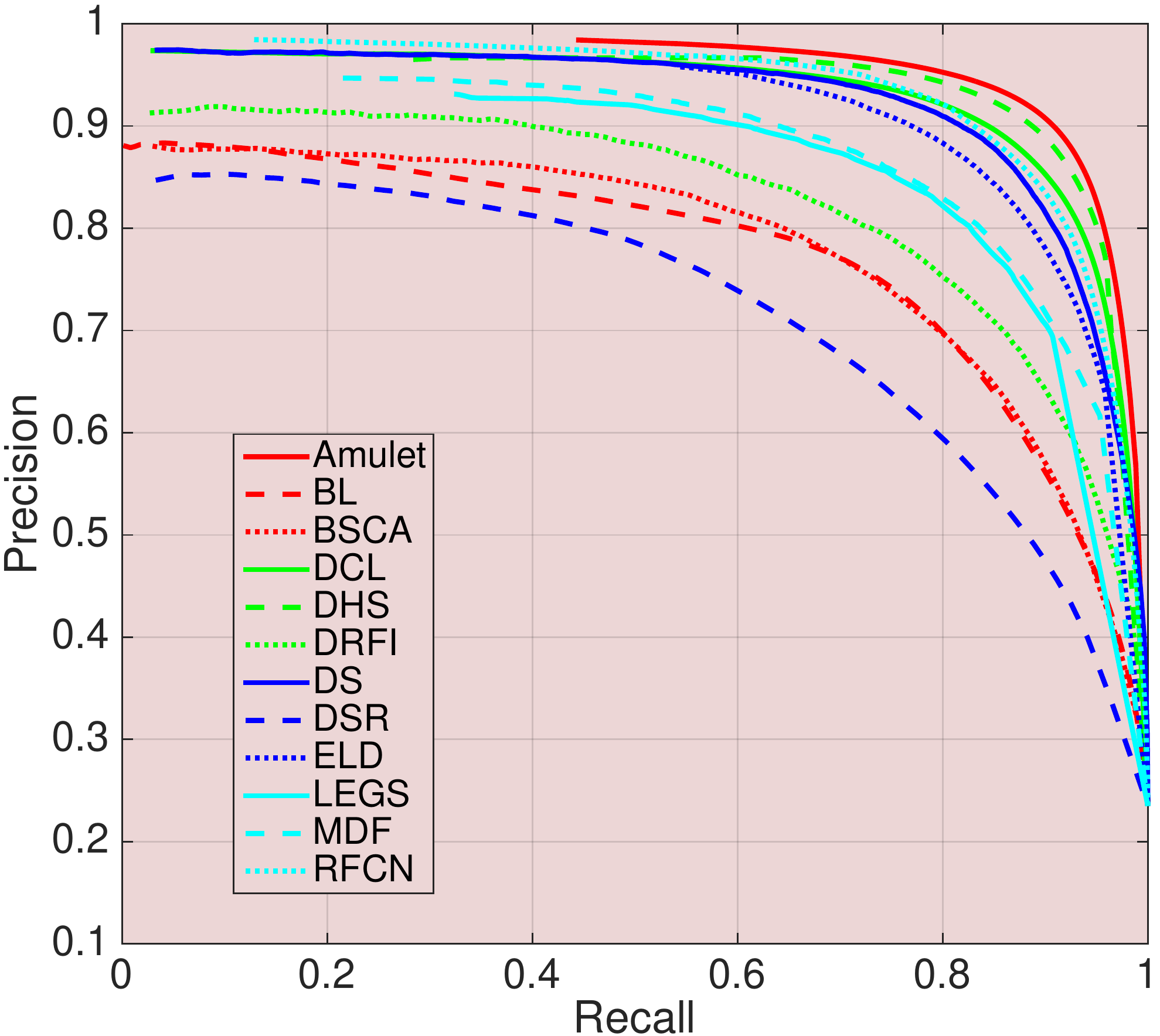} \ &
\includegraphics[width=0.245\linewidth,height=3.85cm]{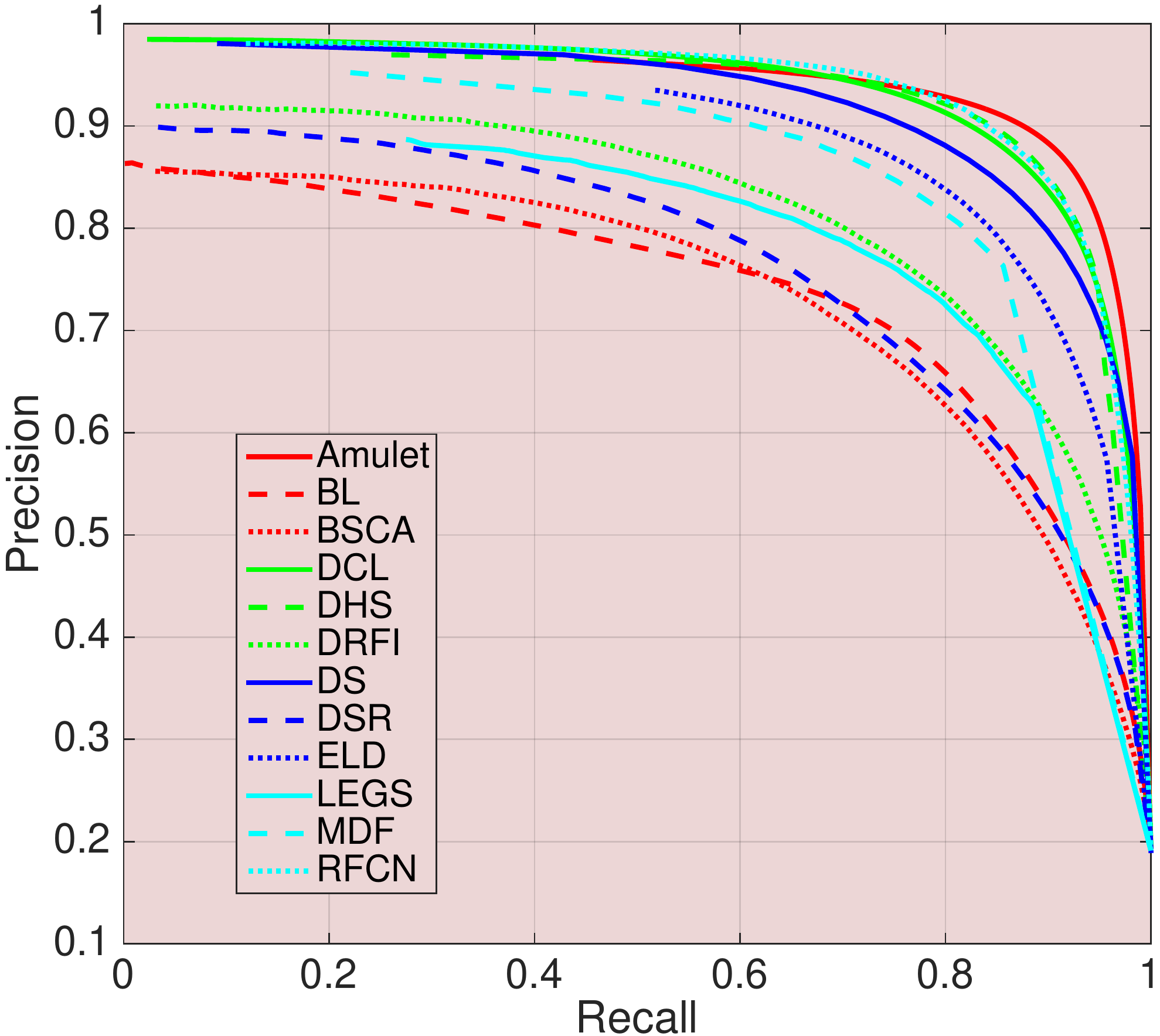} \ &
\includegraphics[width=0.245\linewidth,height=3.85cm]{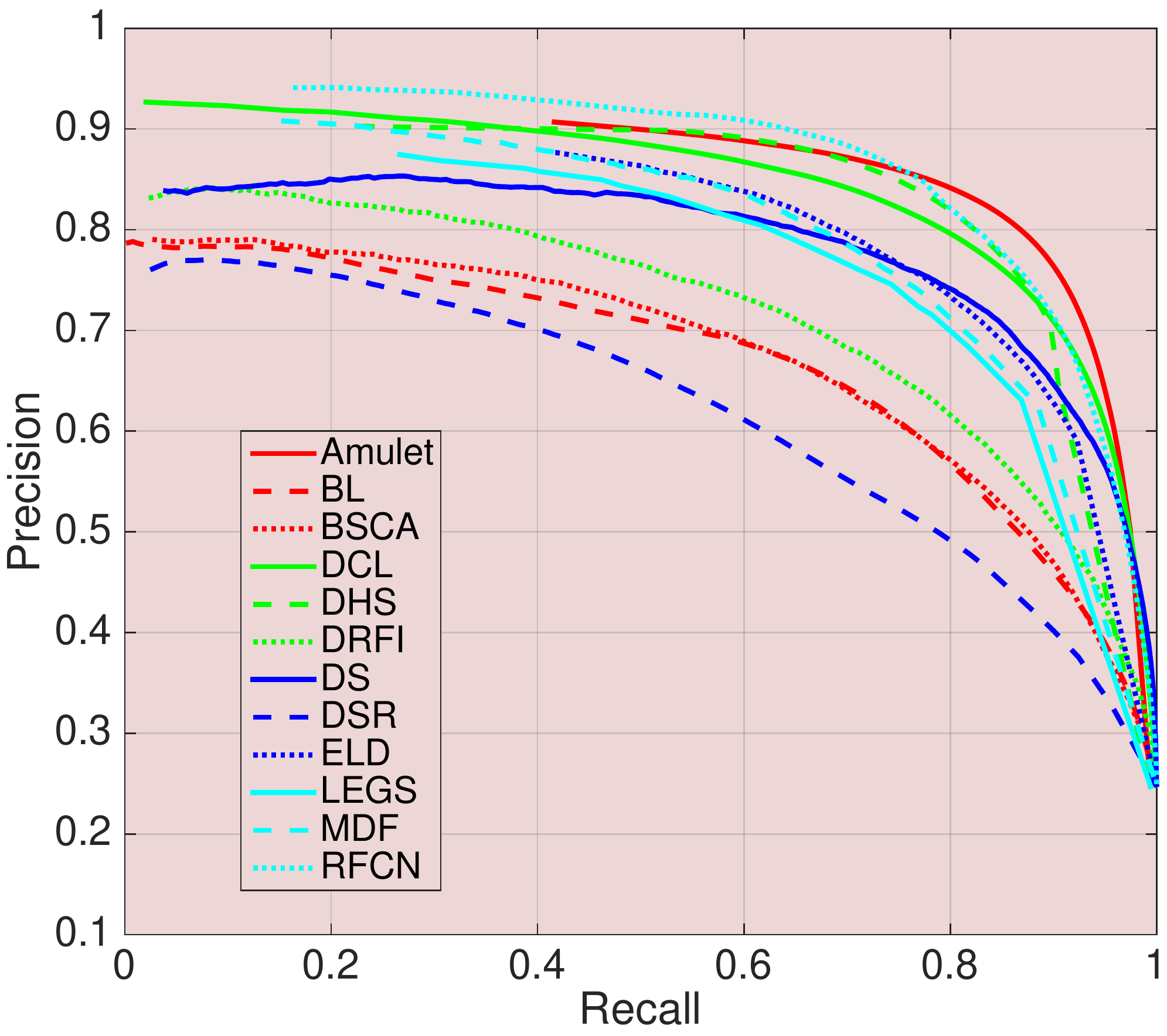} \ \\
 {\small(a) DUTS-TE} & {\small(b) ECSSD} & {\small(c) HKU-IS} & {\small(d) PASCAL-S}\\ \\
\end{tabular}
\vspace{-6mm}
\caption{The PR curves of the proposed algorithm and other state-of-the-art methods.
\label{fig:PR-curve}}
\end{center}
\vspace{-10mm}
\end{figure*}

\textbf{Quantitative Evaluation}. As shown in Tab.~\ref{table:fauc} and Fig.~\ref{fig:PR-curve}, the \textbf{Amulet} model can largely outperform other compared counterparts across all the datasets in terms of near all evaluation metrics, which convincingly demonstrates the effectiveness of the proposed method.
Results on the SED dataset and PR curves on the DUT-OMRON, SED and SOD datasets appear in the supplemental material due to the limitation of space.
From the results, we have other fundamental observations: (1) Our model improves the F-measure with a considerable margin on most of datasets, especially on large-scale datasets, such as DUTS-TE, ECSSD, HKU-IS.
And at the same time, our model generally decreases the MAE.
This indicates that our model is more convinced of the predicted regions and provides more accurate saliency maps.
(2) Although only trained on the MSRA10K dataset, our model significantly outperforms other algorithms that pre-trained on specific saliency datasets, such as LEGS and RFCN on PASCAL-S, MDF on HKU-IS.
The superior performance confirms that our model have good generalization abilities on other large-scale datasets.
(3) Our method is inferior to DHS on several datasets. However, these datasets are relatively small compared to the era of deep learning.
%, especially on complex datasets.
%Our model also significantly outperforms the FCN adapted from a model originally designed
%for semantic segmentation [30] because we explicitly perform deep contrast learning, which is critical for saliencydetection.

\begin{figure*}
\centering
\begin{tabular}{@{}c@{}c@{}c@{}c@{}c@{}c@{}c@{}c@{}c@{}c@{}c}
%%%%  contrast
%\vspace{-0.7mm}
%\includegraphics[width=0.09\linewidth,height=1.25cm]{figs/Amulet_images/0011.jpg} \ &
%\includegraphics[width=0.09\linewidth,height=1.25cm]{figs/Amulet_images/gt0011.png} \ &
%\includegraphics[width=0.09\linewidth,height=1.25cm]{figs/Amulet_images/96/0011.png} \ &
%\includegraphics[width=0.09\linewidth,height=1.25cm]{figs/Amulet_images/0011_RFCN.jpg} \ &
%\includegraphics[width=0.09\linewidth,height=1.25cm]{figs/Amulet_images/0011_DCL.png} \ &
%\includegraphics[width=0.09\linewidth,height=1.25cm]{figs/Amulet_images/0011_DS.png} \ &
%\includegraphics[width=0.09\linewidth,height=1.25cm]{figs/Amulet_images/0011_LEGS.png} \ &
%\includegraphics[width=0.09\linewidth,height=1.25cm]{figs/Amulet_images/0011_MDF.png} \ &
%\includegraphics[width=0.09\linewidth,height=1.25cm]{figs/Amulet_images/0011_ELD.png} \ &
%\includegraphics[width=0.09\linewidth,height=1.25cm]{figs/Amulet_images/0011_DRFI.png} \ \\
\vspace{-1mm}
\includegraphics[width=0.085\linewidth,height=1.25cm]{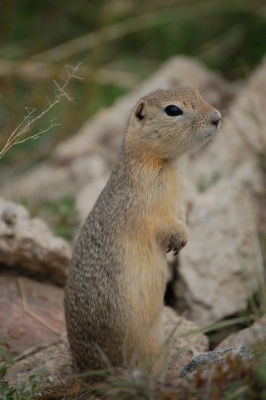} \ &
\includegraphics[width=0.085\linewidth,height=1.25cm]{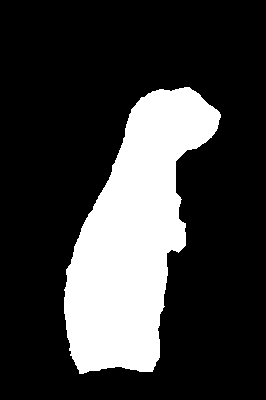} \ &
\includegraphics[width=0.085\linewidth,height=1.25cm]{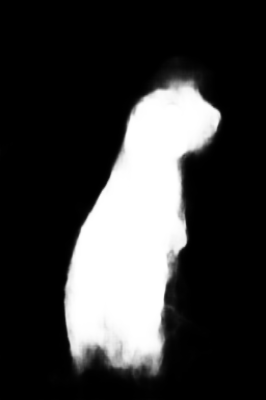} \ &
\includegraphics[width=0.085\linewidth,height=1.25cm]{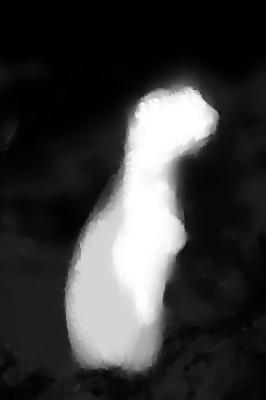} \ &
\includegraphics[width=0.085\linewidth,height=1.25cm]{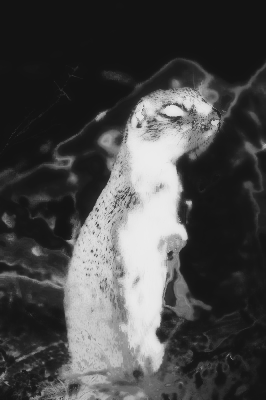} \ &
\includegraphics[width=0.085\linewidth,height=1.25cm]{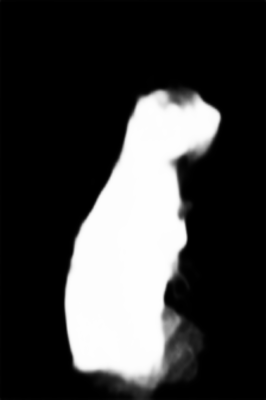} \ &
\includegraphics[width=0.085\linewidth,height=1.25cm]{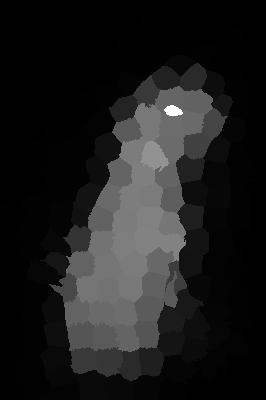} \ &
\includegraphics[width=0.085\linewidth,height=1.25cm]{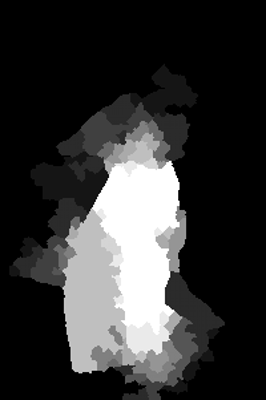} \ &
\includegraphics[width=0.085\linewidth,height=1.25cm]{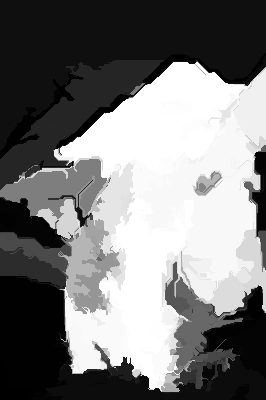} \ &
\includegraphics[width=0.085\linewidth,height=1.25cm]{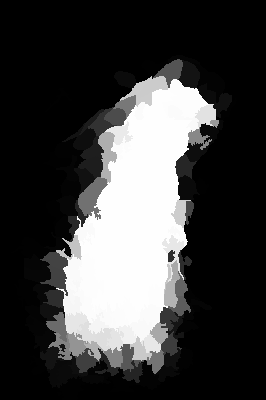} \ &
\includegraphics[width=0.085\linewidth,height=1.25cm]{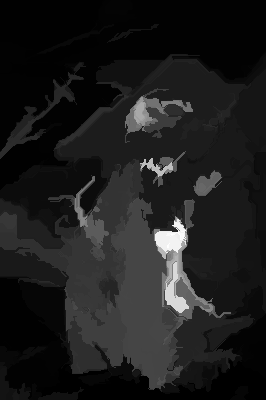} \ \\
\vspace{-1mm}
\includegraphics[width=0.085\linewidth,height=1.25cm]{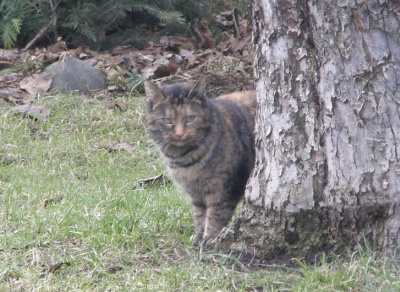} \ &
\includegraphics[width=0.085\linewidth,height=1.25cm]{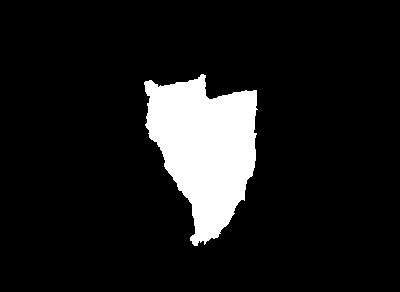} \ &
\includegraphics[width=0.085\linewidth,height=1.25cm]{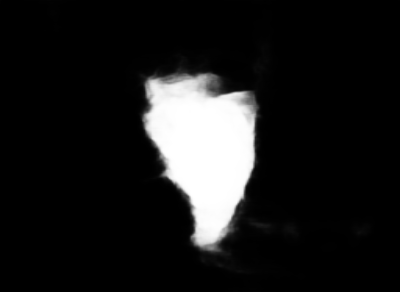} \ &
\includegraphics[width=0.085\linewidth,height=1.25cm]{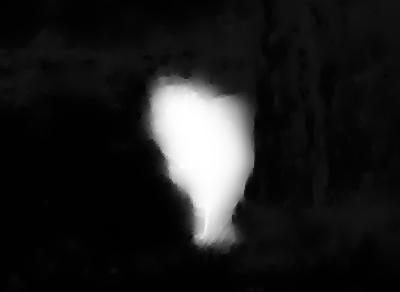} \ &
\includegraphics[width=0.085\linewidth,height=1.25cm]{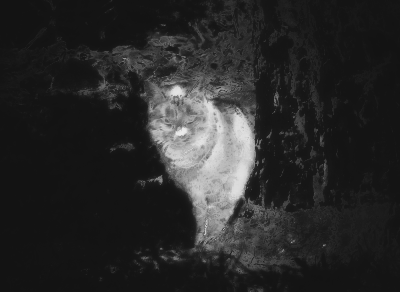} \ &
\includegraphics[width=0.085\linewidth,height=1.25cm]{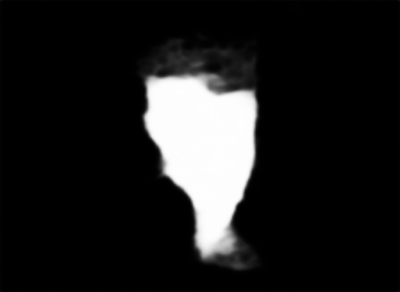} \ &
\includegraphics[width=0.085\linewidth,height=1.25cm]{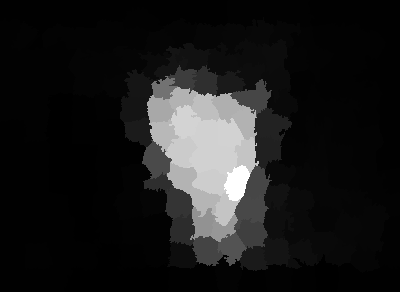} \ &
\includegraphics[width=0.085\linewidth,height=1.25cm]{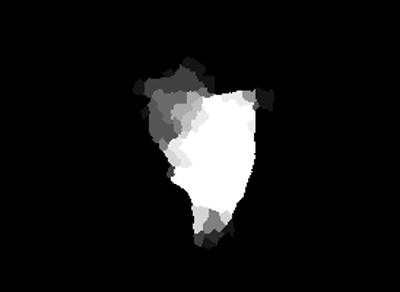} \ &
\includegraphics[width=0.085\linewidth,height=1.25cm]{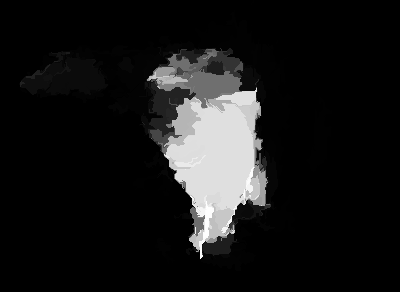} \ &
\includegraphics[width=0.085\linewidth,height=1.25cm]{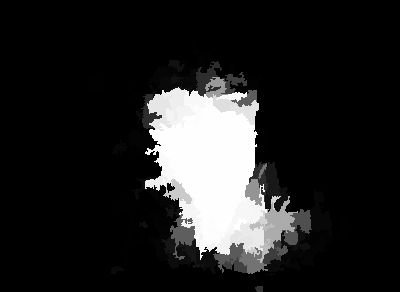} \ &
\includegraphics[width=0.085\linewidth,height=1.25cm]{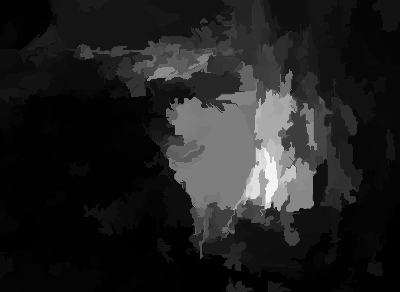} \ \\
%\vspace{-0.7mm}
%\includegraphics[width=0.09\linewidth,height=1.25cm]{figs/Amulet_images/0125.jpg} \ &
%\includegraphics[width=0.09\linewidth,height=1.25cm]{figs/Amulet_images/gt0125.png} \ &
%\includegraphics[width=0.09\linewidth,height=1.25cm]{figs/Amulet_images/96/0125.png} \ &
%\includegraphics[width=0.09\linewidth,height=1.25cm]{figs/Amulet_images/0125_RFCN.png} \ &
%\includegraphics[width=0.09\linewidth,height=1.25cm]{figs/Amulet_images/0125_DCL.png} \ &
%\includegraphics[width=0.09\linewidth,height=1.25cm]{figs/Amulet_images/0125_DS.png} \ &
%\includegraphics[width=0.09\linewidth,height=1.25cm]{figs/Amulet_images/0125_LEGS.png} \ &
%\includegraphics[width=0.09\linewidth,height=1.25cm]{figs/Amulet_images/0125_MDF.png} \ &
%\includegraphics[width=0.09\linewidth,height=1.25cm]{figs/Amulet_images/0125_ELD.png} \ &
%\includegraphics[width=0.09\linewidth,height=1.25cm]{figs/Amulet_images/0125_DRFI.png} \ \\
%%%%  image boundary
\vspace{-1mm}
\includegraphics[width=0.085\linewidth,height=1.25cm]{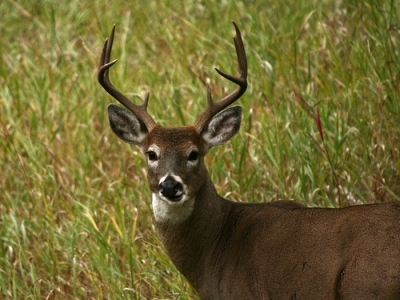} \ &
\includegraphics[width=0.085\linewidth,height=1.25cm]{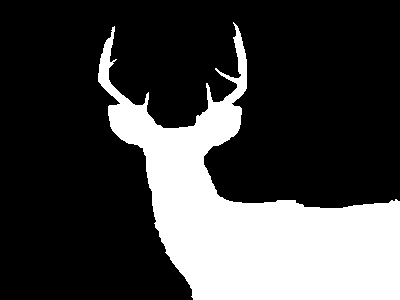} \ &
\includegraphics[width=0.085\linewidth,height=1.25cm]{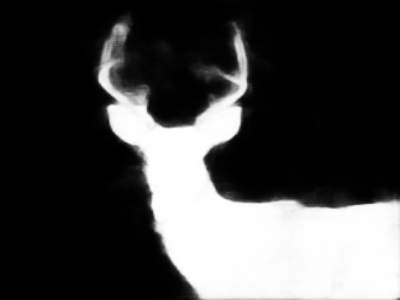} \ &
\includegraphics[width=0.085\linewidth,height=1.25cm]{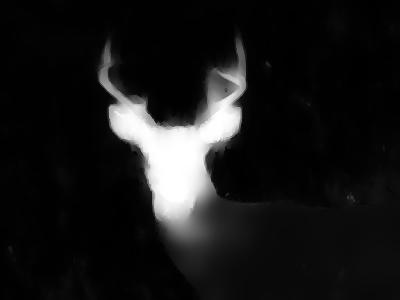} \ &
\includegraphics[width=0.085\linewidth,height=1.25cm]{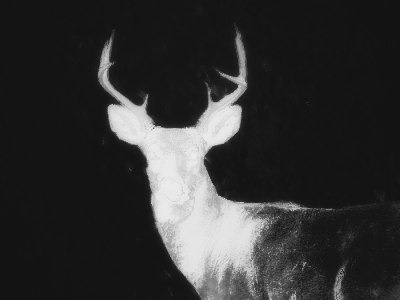} \ &
\includegraphics[width=0.085\linewidth,height=1.25cm]{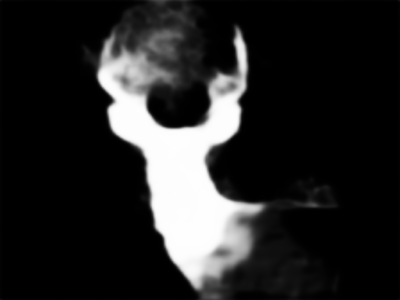} \ &
\includegraphics[width=0.085\linewidth,height=1.25cm]{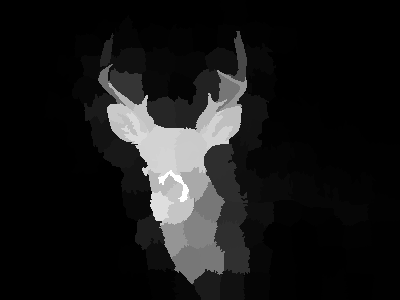} \ &
\includegraphics[width=0.085\linewidth,height=1.25cm]{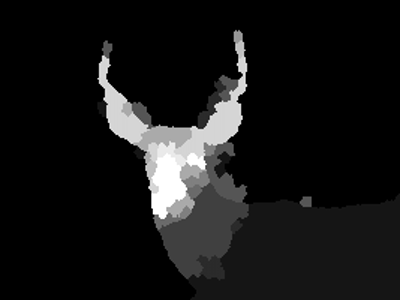} \ &
\includegraphics[width=0.085\linewidth,height=1.25cm]{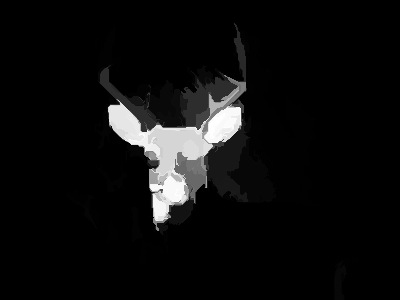} \ &
\includegraphics[width=0.085\linewidth,height=1.25cm]{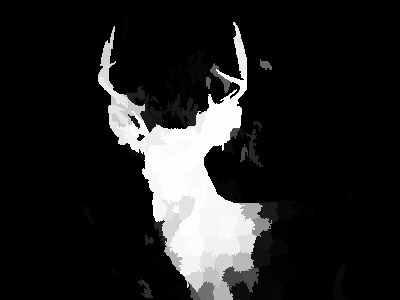} \ &
\includegraphics[width=0.085\linewidth,height=1.25cm]{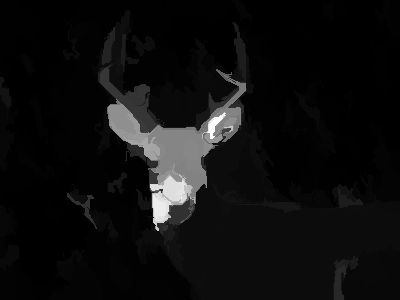} \ \\
\vspace{-1mm}
\includegraphics[width=0.085\linewidth,height=1.25cm]{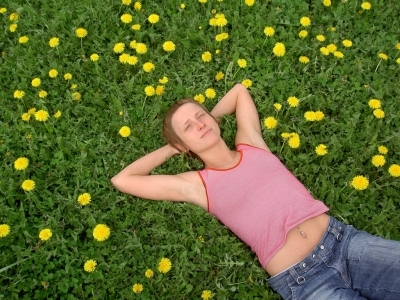} \ &
\includegraphics[width=0.085\linewidth,height=1.25cm]{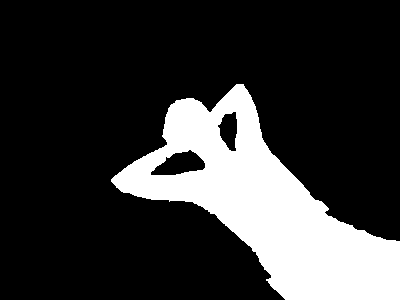} \ &
\includegraphics[width=0.085\linewidth,height=1.25cm]{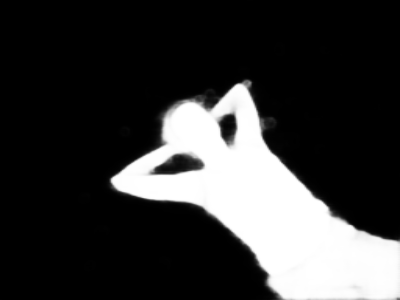} \ &
\includegraphics[width=0.085\linewidth,height=1.25cm]{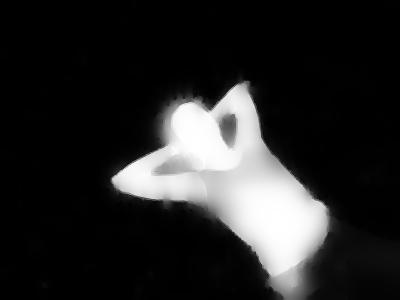} \ &
\includegraphics[width=0.085\linewidth,height=1.25cm]{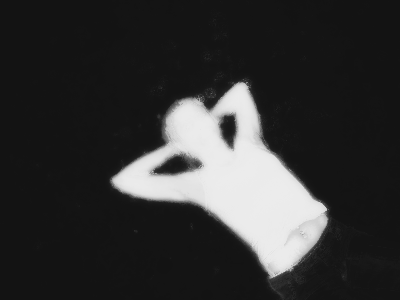} \ &
\includegraphics[width=0.085\linewidth,height=1.25cm]{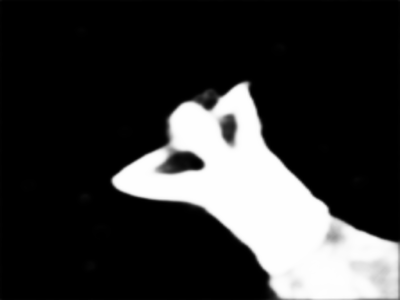} \ &
\includegraphics[width=0.085\linewidth,height=1.25cm]{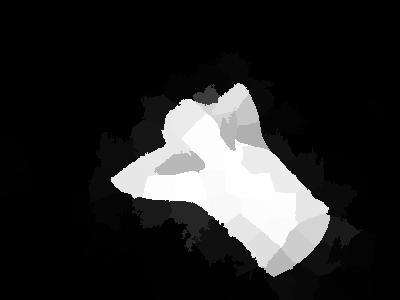} \ &
\includegraphics[width=0.085\linewidth,height=1.25cm]{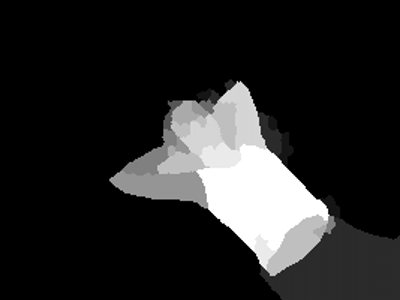} \ &
\includegraphics[width=0.085\linewidth,height=1.25cm]{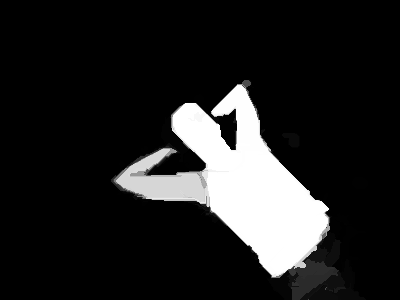} \ &
\includegraphics[width=0.085\linewidth,height=1.25cm]{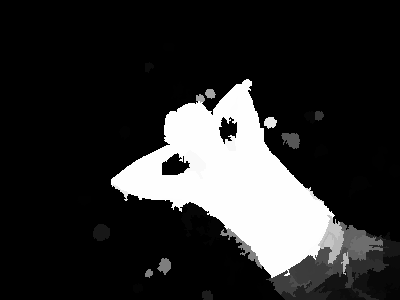} \ &
\includegraphics[width=0.085\linewidth,height=1.25cm]{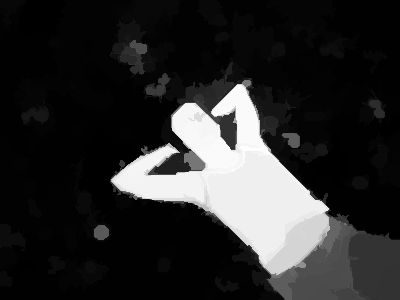} \ \\
\vspace{-1mm}
\includegraphics[width=0.085\linewidth,height=1.25cm]{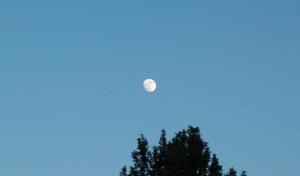} \ &
\includegraphics[width=0.085\linewidth,height=1.25cm]{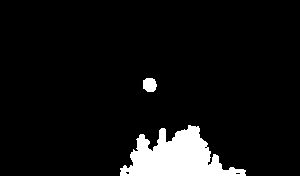} \ &
\includegraphics[width=0.085\linewidth,height=1.25cm]{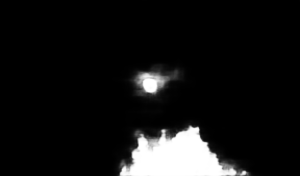} \ &
\includegraphics[width=0.085\linewidth,height=1.25cm]{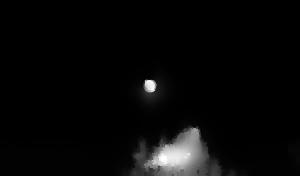} \ &
\includegraphics[width=0.085\linewidth,height=1.25cm]{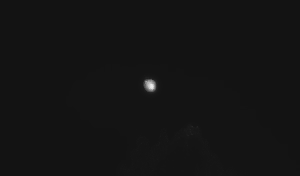} \ &
\includegraphics[width=0.085\linewidth,height=1.25cm]{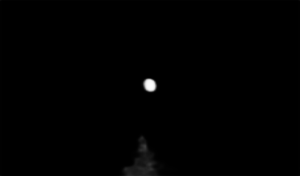} \ &
\includegraphics[width=0.085\linewidth,height=1.25cm]{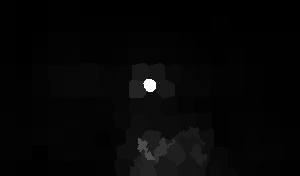} \ &
\includegraphics[width=0.085\linewidth,height=1.25cm]{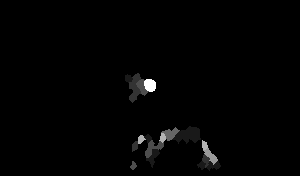} \ &
\includegraphics[width=0.085\linewidth,height=1.25cm]{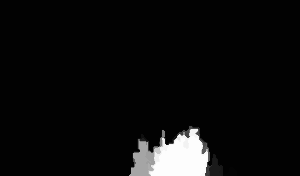} \ &
\includegraphics[width=0.085\linewidth,height=1.25cm]{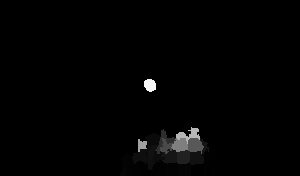} \ &
\includegraphics[width=0.085\linewidth,height=1.25cm]{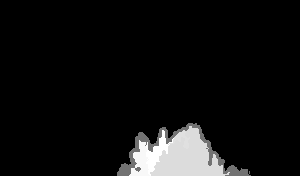} \ \\
%\vspace{-0.5mm}
%\includegraphics[width=0.09\linewidth,height=1.25cm]{figs/Amulet_images/0348.jpg} \ &
%\includegraphics[width=0.09\linewidth,height=1.25cm]{figs/Amulet_images/gt0348.png} \ &
%\includegraphics[width=0.09\linewidth,height=1.25cm]{figs/Amulet_images/96/0348.png} \ &
%\includegraphics[width=0.09\linewidth,height=1.25cm]{figs/Amulet_images/0348_RFCN.png} \ &
%\includegraphics[width=0.09\linewidth,height=1.25cm]{figs/Amulet_images/0348_DCL.png} \ &
%\includegraphics[width=0.09\linewidth,height=1.25cm]{figs/Amulet_images/0348_DS.png} \ &
%\includegraphics[width=0.09\linewidth,height=1.25cm]{figs/Amulet_images/0348_LEGS.png} \ &
%\includegraphics[width=0.09\linewidth,height=1.25cm]{figs/Amulet_images/0348_MDF.png} \ &
%\includegraphics[width=0.09\linewidth,height=1.25cm]{figs/Amulet_images/0348_ELD.png} \ &
%\includegraphics[width=0.09\linewidth,height=1.25cm]{figs/Amulet_images/0348_DRFI.png} \ \\
\vspace{-1mm}
\includegraphics[width=0.085\linewidth,height=1.25cm]{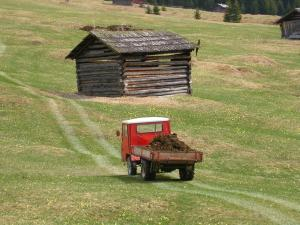} \ &
\includegraphics[width=0.085\linewidth,height=1.25cm]{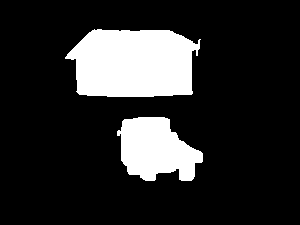} \ &
\includegraphics[width=0.085\linewidth,height=1.25cm]{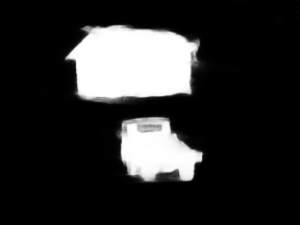} \ &
\includegraphics[width=0.085\linewidth,height=1.25cm]{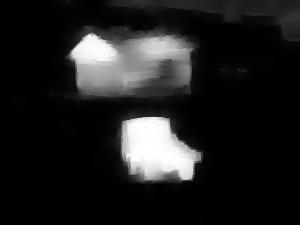} \ &
\includegraphics[width=0.085\linewidth,height=1.25cm]{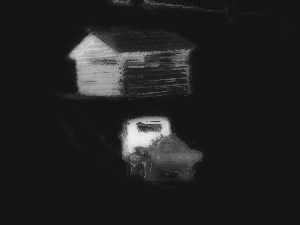} \ &
\includegraphics[width=0.085\linewidth,height=1.25cm]{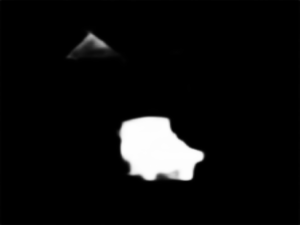} \ &
\includegraphics[width=0.085\linewidth,height=1.25cm]{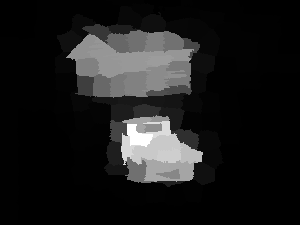} \ &
\includegraphics[width=0.085\linewidth,height=1.25cm]{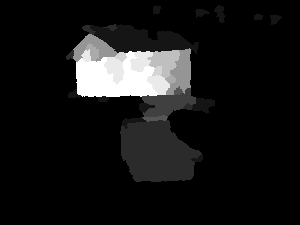} \ &
\includegraphics[width=0.085\linewidth,height=1.25cm]{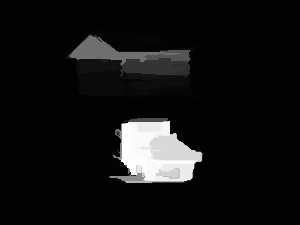} \ &
\includegraphics[width=0.085\linewidth,height=1.25cm]{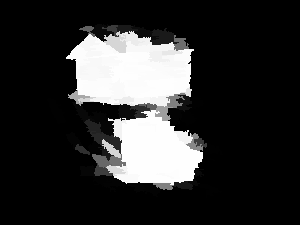} \ &
\includegraphics[width=0.085\linewidth,height=1.25cm]{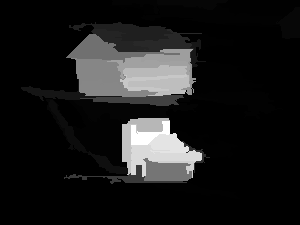} \ \\
{\small (a)} & {\small(b)} & {\small(c)} & {\small(d)} & {\small(e)}& {\small(f)}& {\small(g)}
& {\small(h)}& {\small(i)}& {\small(j)}& {\small(k)} \\
\end{tabular}
\caption{Comparison of saliency maps. (a) Input images; (b) Ground truth; (c) Our method; (d) RFCN; (e) DCL; (f) DHS; (g) DS; (h) LEGS; (i) MDF; (j) ELD; (k) DRFI. The top four row and bottom two row images are from the ECSSD and SED dataset, respectively.
\label{fig:map_examples}}
\end{figure*}

\textbf{Qualitative Evaluation}. Fig.~\ref{fig:map_examples} provides a visual comparison of our approach and other methods.
It can be seen that our method generates more accurate saliency maps in various challenging cases, \eg, low contrast between the objects and backgrounds (the first two rows), objects near the image boundary (the 3-4 rows) and multiple disconnected salient objects (the 5-6 rows). What's more, with our BPR component, our saliency maps provide more accurate boundaries of salient objects (the 1, 3, 4, 6 rows).
%%%%%
%%%%%%%%%%%%%%%%%%%%%%%%%%%%%%%%%%%%%%%%%%%%%
\begin{figure*}
\vspace{-3mm}
\centering
\begin{tabular}{@{}c@{}c@{}c@{}c@{}c@{}c@{}c@{}c@{}c@{}}
\vspace{-0.5mm}
\includegraphics[width=0.105\linewidth,height=1.8cm]{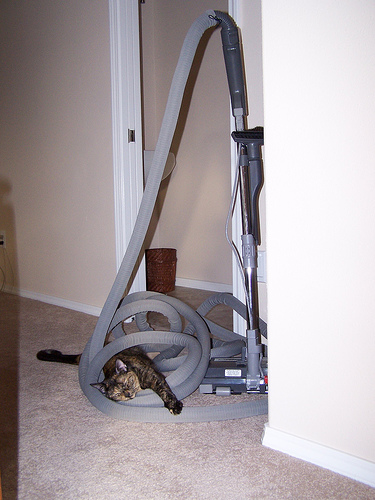} \ &
\includegraphics[width=0.105\linewidth,height=1.8cm]{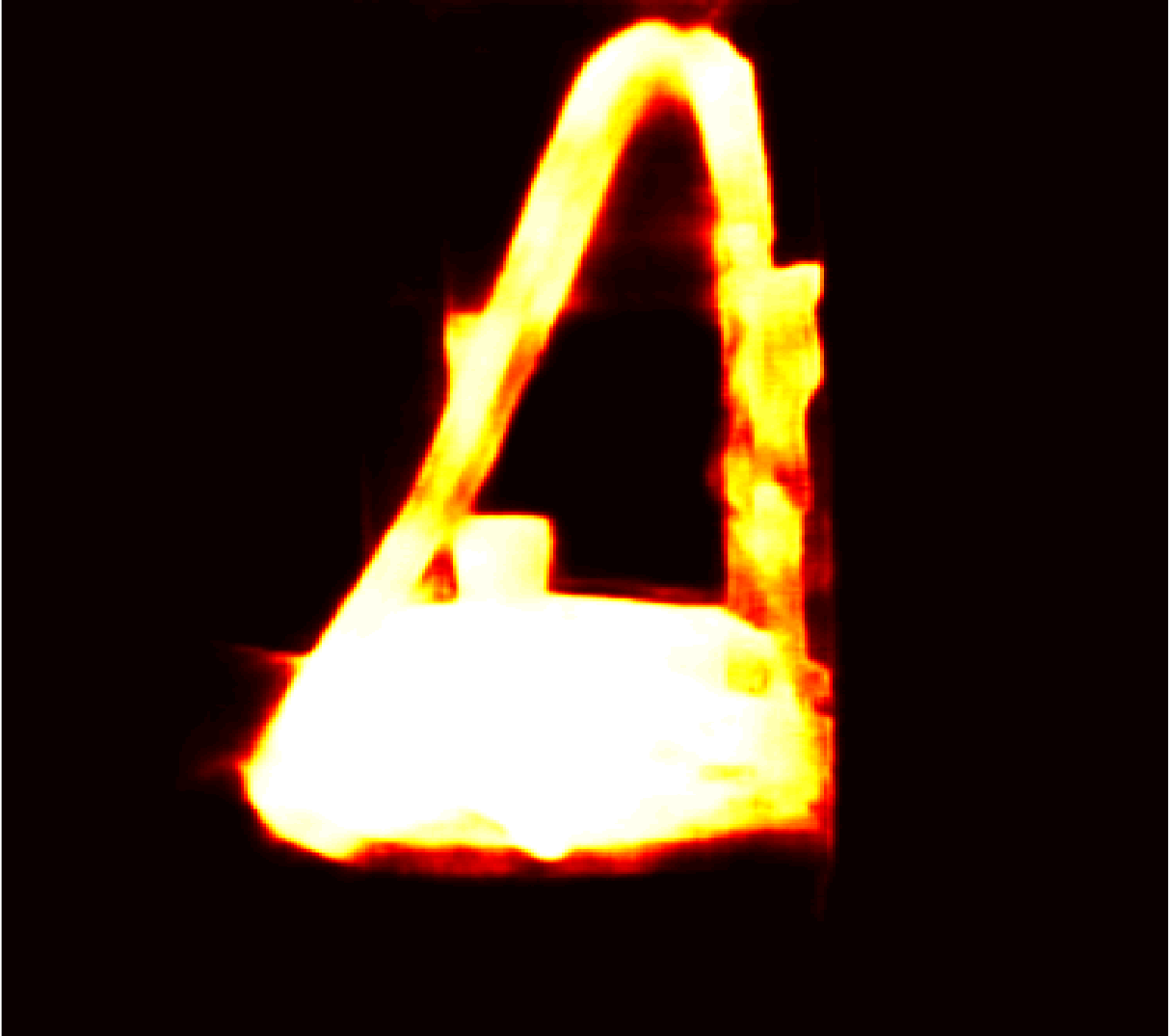} \ &
\includegraphics[width=0.105\linewidth,height=1.8cm]{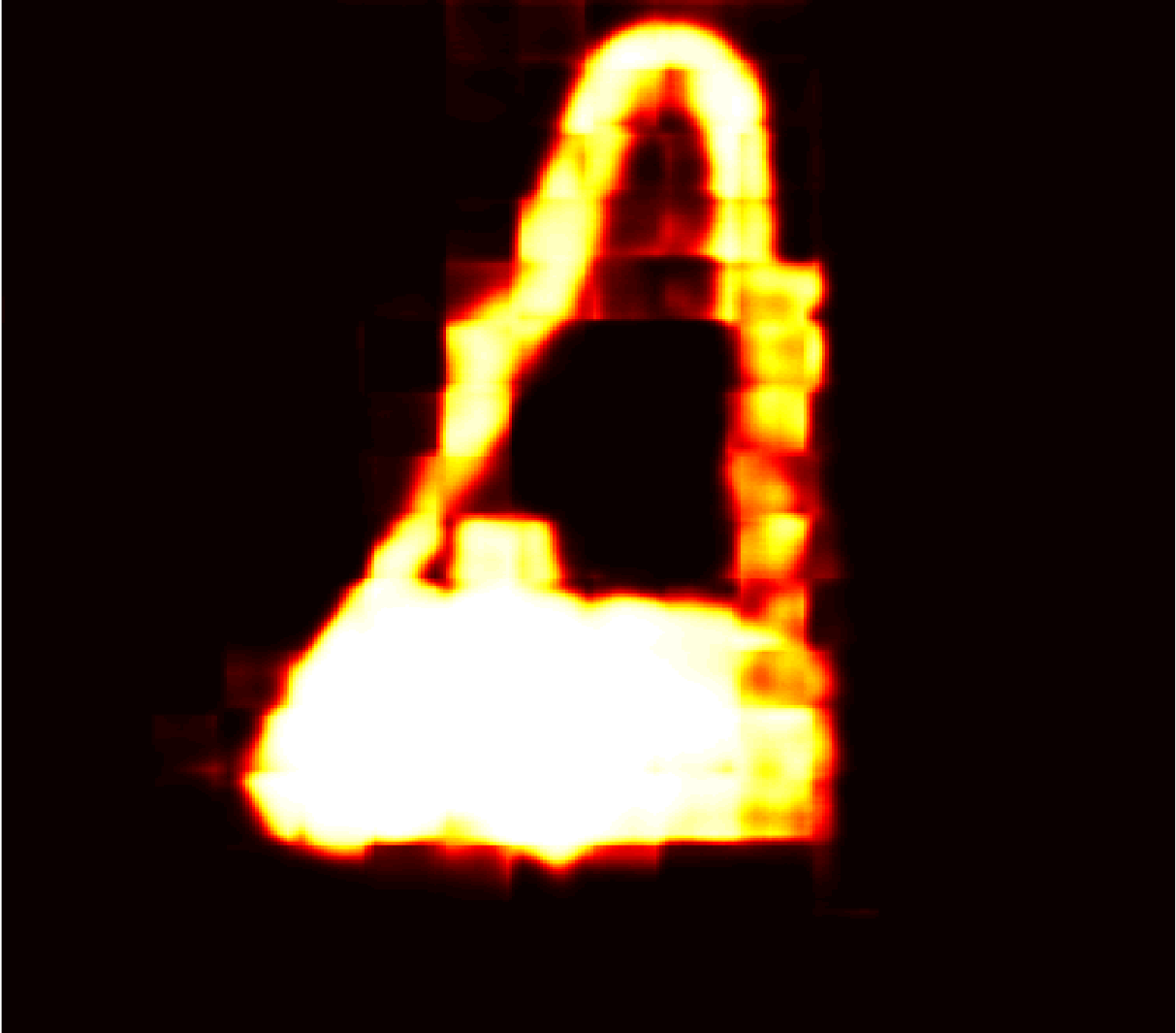} \ &
\includegraphics[width=0.105\linewidth,height=1.8cm]{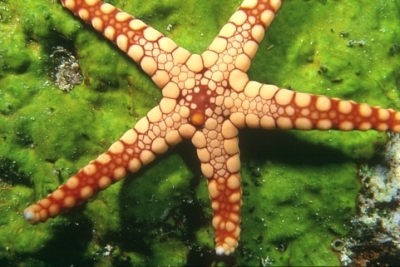} \ &
\includegraphics[width=0.105\linewidth,height=1.8cm]{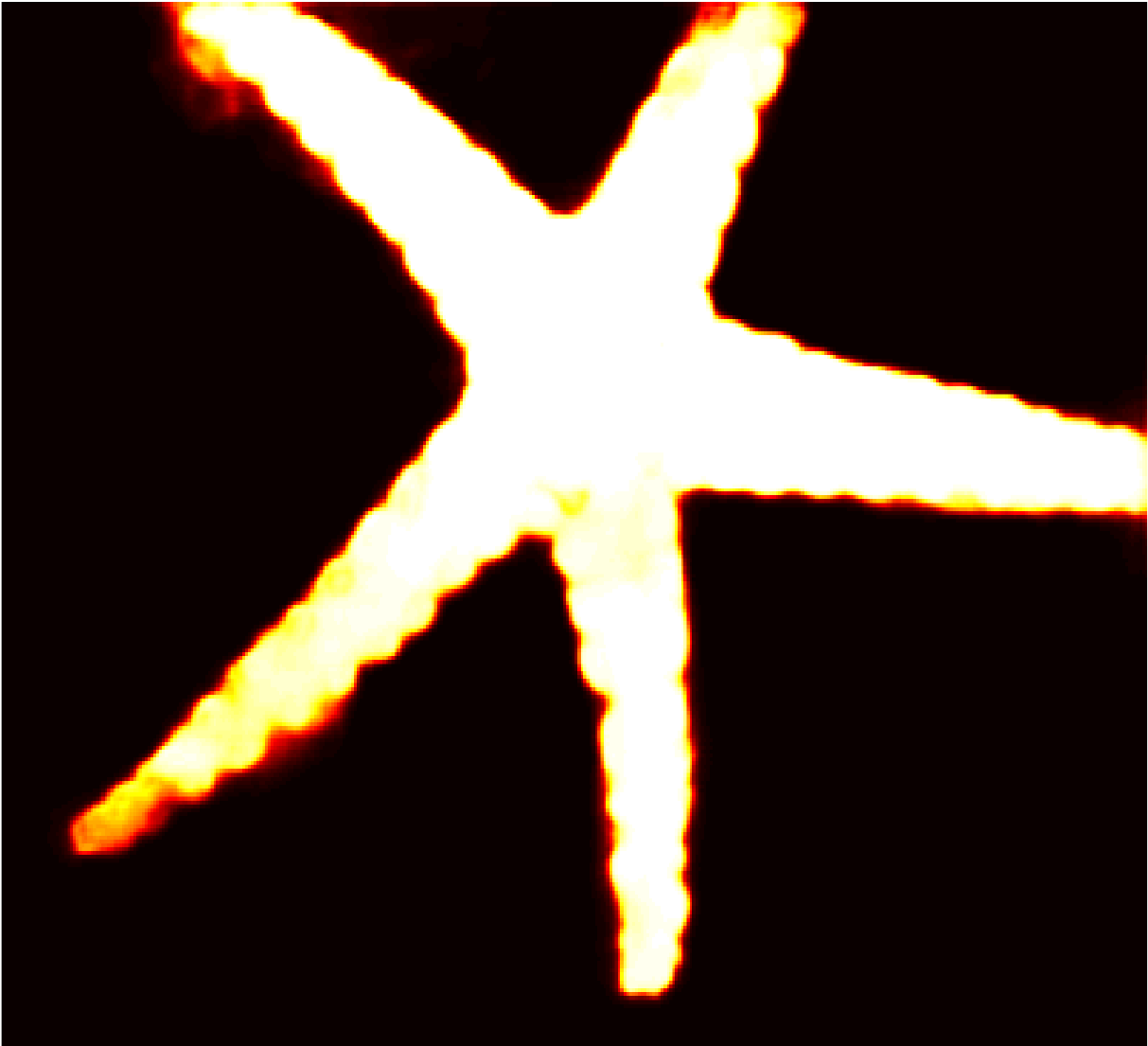} \ &
\includegraphics[width=0.105\linewidth,height=1.8cm]{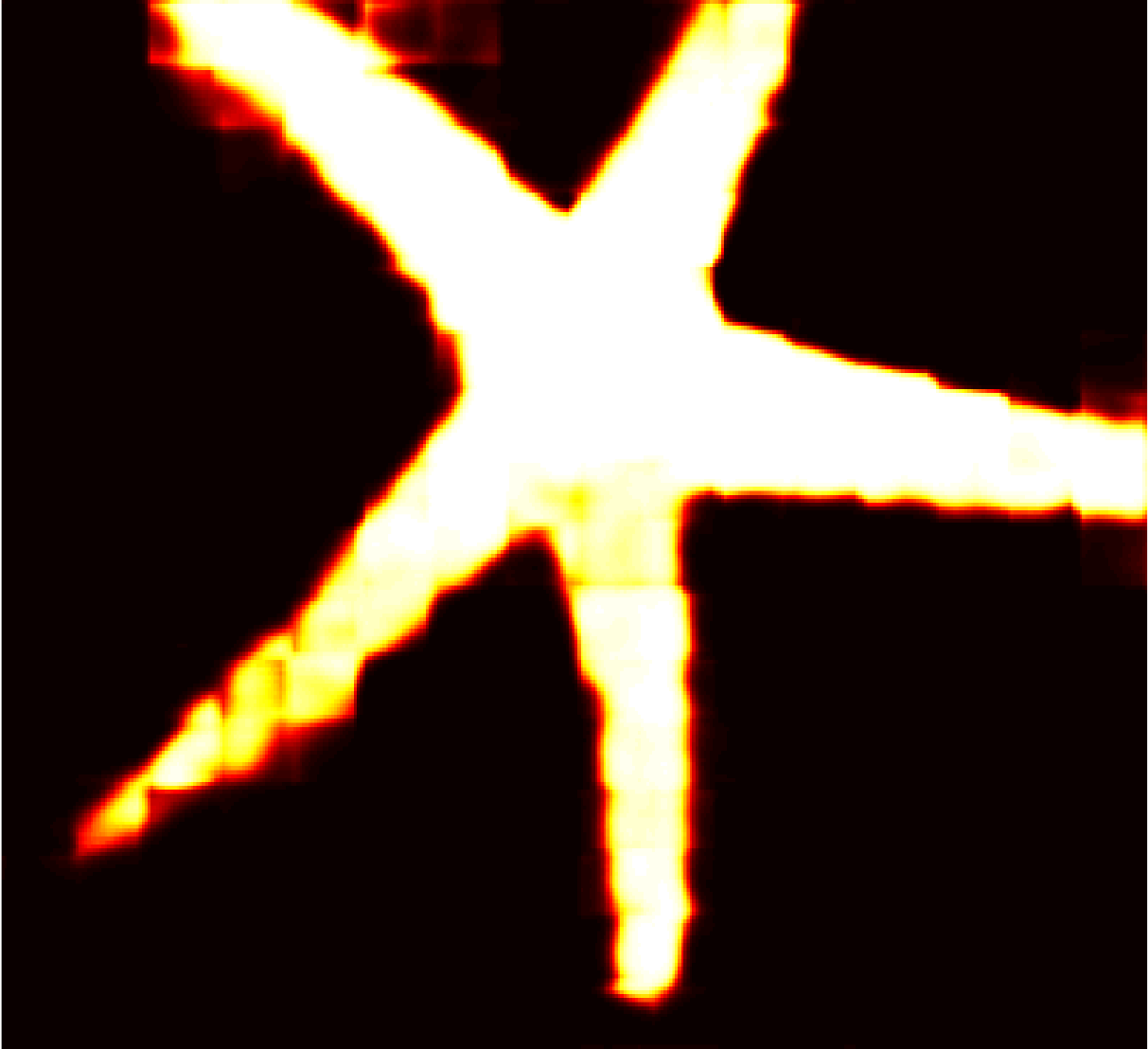} \ &
\includegraphics[width=0.105\linewidth,height=1.8cm]{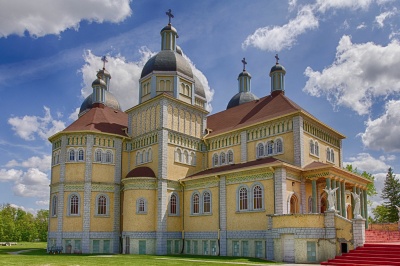} \ &
\includegraphics[width=0.105\linewidth,height=1.8cm]{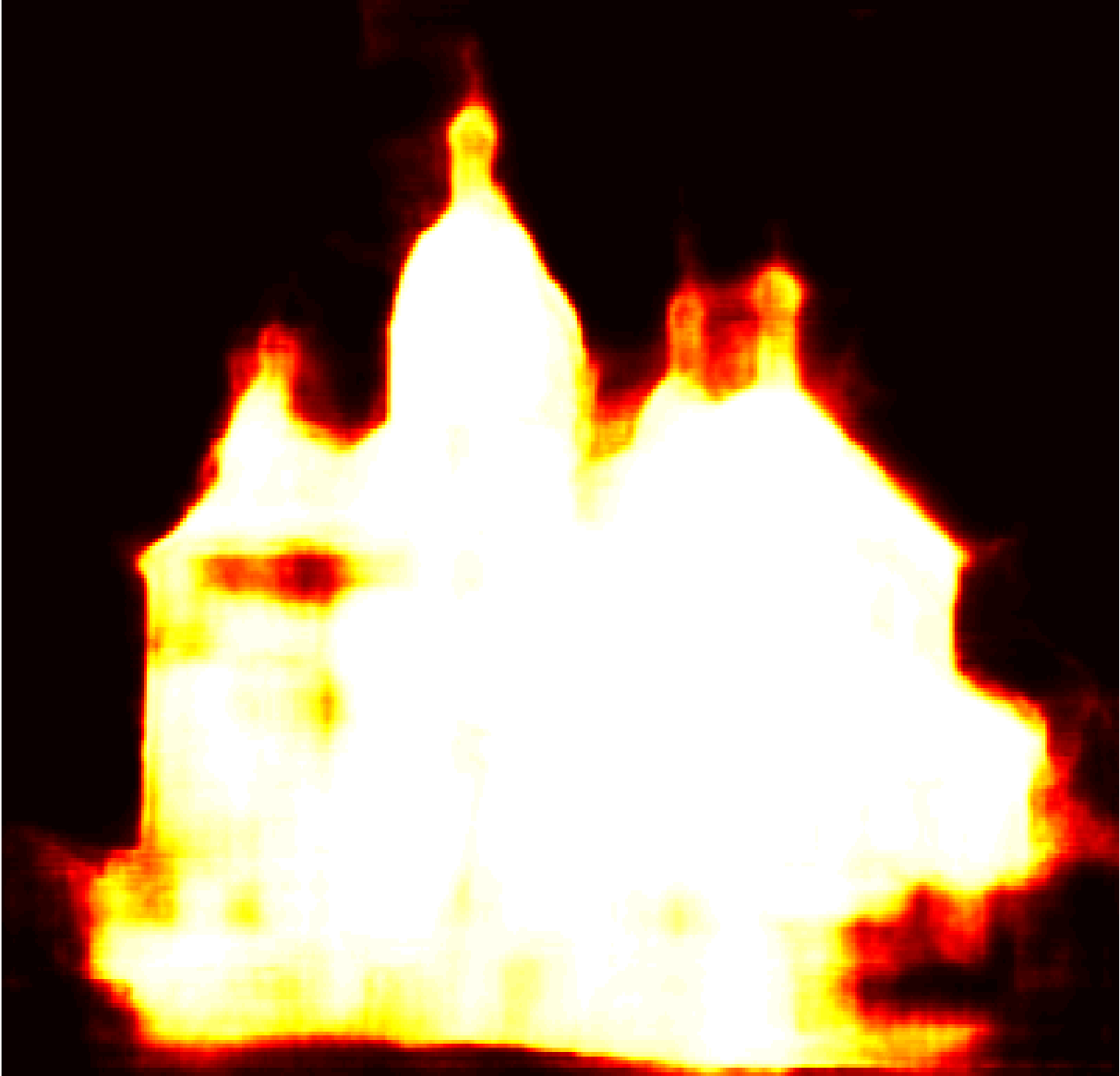} \ &
\includegraphics[width=0.105\linewidth,height=1.8cm]{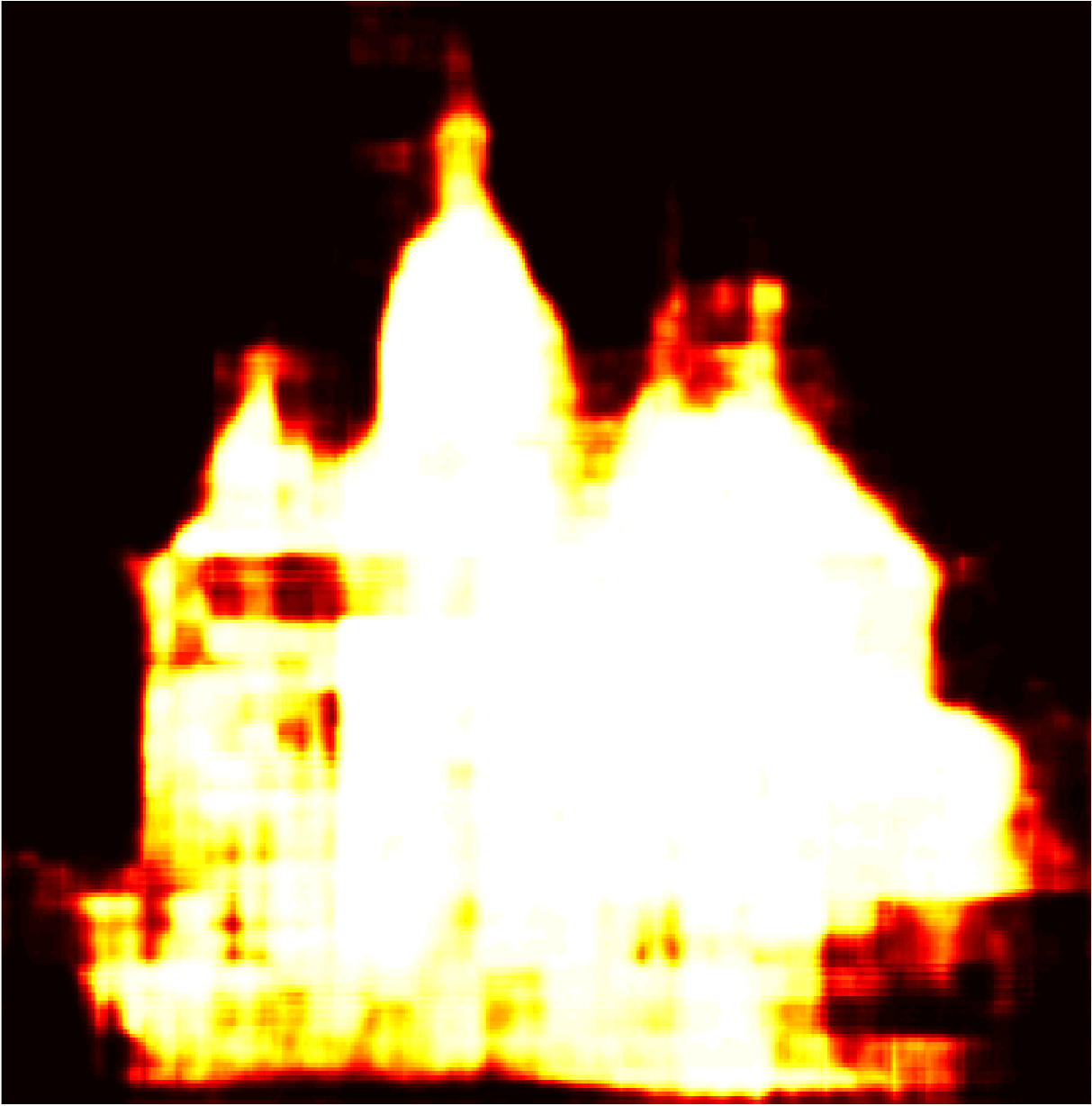} \ \\
{\small(a)} & {\small(b)} & {\small(c)} & {\small(d)} & {\small(e)} & {\small(f)}& {\small(g)} & {\small(h)} & {\small(i)}\\
\end{tabular}
\caption{Visual comparison of the Amulet algorithm with /without BPRs. (a)(d)(g) Input images; (b)(e)(h) Predictions of the \textbf{Amulet}; (c)(f)(i) Predictions of the $\textbf{Amulet}_{BPR^{-}}$. High resolution to see better.
\label{fig:Ablation}}
\vspace{-5mm}
\end{figure*}
\setlength{\tabcolsep}{1.75pt}
\subsection{Ablation Studies}
\textbf{Feature resolution effects.} To verify the importance of resolutions of integrated features, we additionally evaluate several variants of the proposed \textbf{Amulet} model with different scales.
\textbf{Amulet}-$1/n$ denotes the model that takes the integrated features reduced by a factor not larger than $n$, with respect to the input image.
The corresponding performance are also reported in Tab.~\ref{table:fauc}.
The results suggest that features of all levels are helpful for saliency detection, and with the increment of resolutions, our approach gradually achieves better performance.
In addition, even our simplest model (i.e., \textbf{Amulet}-1/16) can achieve better results than most of existing methods.
This fact further verifies the strength of our proposed methods.

\textbf{Boundary refinements.}
To verify the contributions of our proposed BPR, we also implement our proposed approach without BPRs, named $\textbf{Amulet}_{BPR^{-}}$, and report the performance in Tab.~\ref{table:fauc}.
It can be observed that without BPRs, our approach decreases the performance but not too much in F-measure.
But it leads to a large drop in MAE. This indicates that our proposed BPR is capable of detecting and
localizing the boundary of most salient objects, while other methods often fail at this fact.
Several visual examples are illustrated in Fig.~\ref{fig:Ablation}.
\subsection{Comparison with Other Aggregation Methods}
For fair comparison, we perform additional evaluations to verify the detection ability of different aggregation methods.
Specifically, we use the same augmented MSRA10K dataset to train the FCN-8s~\cite{long2015fully}, Hypercolumn (HC)~\cite{hariharan2015hypercolumns}, SegNet (SN)~\cite{segnet}, DeconvNet(DN)~\cite{noh2015learning} and HED~\cite{xie2015holistically} for saliency detection task.
All compared methods are based on the same VGG-16 model pre-trained on the ImageNet classification task~\cite{simonyan2014very}.
We drop the unnecessary components in each model and only focus on the feature aggregation part.
For our model, we use the simplest model (i.e., \textbf{Amulet}-1/16) without BPRs.
For each method, we find the optimal parameters to achieve its' best results.
The performance on the ECSSD dataset is listed in Tab.~\ref{table:aggregation}. As can be seen from Tab.~\ref{table:aggregation}, with the aggregation of multi-level features, our approach achieves better performance.
\begin{table}
\begin{center}
\begin{tabular}{|c|c|c|c|c|c|c|}
\hline
Methods &\textbf{FCN-8s}&\textbf{HC}&\textbf{SN}&\textbf{DN}&\textbf{HED}&\textbf{Ours}\\
\hline
$F_\beta$&0.8116&0.8187& 0.8145 & 0.8264 &0.8321&0.8521        \\
\hline
$MAE$    &0.1343&0.1193& 0.0947& 0.1435& 0.1022&0.0662       \\
\hline
\end{tabular}
\end{center}
\vspace{-3mm}
\caption{The performance of different aggregations on ECSSD dataset. Other datasets have the similar performance trend.}
\label{table:aggregation}
\vspace{-6mm}
\end{table}
%----------------------------------------------------------------
\section{Conclusion}

In this paper, we propose a generic aggregating multi-level convolutional feature framework for salient object detection.
Our framework can integrate multi-level feature maps into multiple resolutions, learn to combine feature maps, and predict saliency maps with the integrated features.
In addition, edge-aware maps and high-level predictions are embedded into the framework.
%
%By aggregating multi-level convolutional features, the proposed saliency model provides accurate salient object labeling.
%
Experiments demonstrate that our method performs favorably against state-of-the-art approaches in saliency detection.
%-------------------------------------------------------------------------
\vspace{-2mm}
{\small {\flushleft\textbf{Acknowledgment}}.
This paper is supported by the Natural Science Foundation of China \#61472060, \#61502070 and \#61528101.}
\vspace{-5mm}
{\small
\bibliographystyle{ieee}
\bibliography{egbib}

\begin{thebibliography}{10}\itemsep=-1pt

\bibitem{segnet}
V.~Badrinarayanan, A.~Kendall, and R.~Cipolla.
\newblock Segnet: A deep convolutional encoder-decoder architecture for image
  segmentation.
\newblock {\em arXiv preprint arXiv:1511.00561}, 2015.

\bibitem{borj2015salient}
A.~Borji.
\newblock What is a salient object? a dataset and a baseline model for salient
  object detection.
\newblock {\em IEEE TIP}, 24(2):742--756, 2015.

\bibitem{borji2015salient}
A.~Borji, M.-M. Cheng, H.~Jiang, and J.~Li.
\newblock Salient object detection: A benchmark.
\newblock {\em IEEE TIP}, 24(12):5706--5722, 2015.

\bibitem{borji2012adaptive}
A.~Borji, S.~Frintrop, D.~N. Sihite, and L.~Itti.
\newblock Adaptive object tracking by learning background context.
\newblock In {\em CVPRW}, pages 23--30. IEEE, 2012.

\bibitem{ChengPAMI}
M.-M. Cheng, N.~J. Mitra, X.~Huang, P.~H.~S. Torr, and S.-M. Hu.
\newblock Global contrast based salient region detection.
\newblock {\em IEEE TPAMI}, 37(3):569--582, 2015.

\bibitem{dai2016instance}
J.~Dai, K.~He, Y.~Li, S.~Ren, and J.~Sun.
\newblock Instance-sensitive fully convolutional networks.
\newblock In {\em ECCV}, pages 534--549, 2016.

\bibitem{ding2011importance}
Y.~Ding, J.~Xiao, and J.~Yu.
\newblock Importance filtering for image retargeting.
\newblock In {\em CVPR}, pages 89--96, 2011.

\bibitem{donoser2009saliency}
M.~Donoser, M.~Urschler, M.~Hirzer, and H.~Bischof.
\newblock Saliency driven total variation segmentation.
\newblock In {\em ICCV}, pages 817--824, 2009.

\bibitem{Everingham2010ThePV}
M.~Everingham, L.~V. Gool, C.~K.~I. Williams, J.~M. Winn, and A.~Zisserman.
\newblock The pascal visual object classes (voc) challenge.
\newblock {\em IJCV}, 88:303--338, 2010.

\bibitem{Federico12}
Y.~P. Federico~Perazzi, Philipp~Kr{\"a}henb{\"u}hl and A.~Hornung.
\newblock Saliency filters: Contrast based filtering for salient region
  detection.
\newblock In {\em CVPR}, pages 733--740, 2012.

\bibitem{gao20123}
Y.~Gao, M.~Wang, D.~Tao, R.~Ji, and Q.~Dai.
\newblock 3-d object retrieval and recognition with hypergraph analysis.
\newblock {\em IEEE TIP}, 21(9):4290--4303, 2012.

\bibitem{harel07}
J.~Harel, C.~Koch, and P.~Perona.
\newblock Graph-based visual saliency.
\newblock In {\em NIPS}, pages 545--552, 2007.

\bibitem{hariharan2015hypercolumns}
B.~Hariharan, P.~Arbel{\'a}ez, R.~Girshick, and J.~Malik.
\newblock Hypercolumns for object segmentation and fine-grained localization.
\newblock In {\em CVPR}, pages 447--456, 2015.

\bibitem{he2012mobile}
J.~He, J.~Feng, X.~Liu, T.~Cheng, T.-H. Lin, H.~Chung, and S.-F. Chang.
\newblock Mobile product search with bag of hash bits and boundary reranking.
\newblock In {\em CVPR}, pages 3005--3012, 2012.

\bibitem{He2015Delving}
K.~He, X.~Zhang, S.~Ren, and J.~Sun.
\newblock Delving deep into rectifiers: Surpassing human-level performance on
  imagenet classification.
\newblock In {\em ICCV}, pages 1026--1034, 2015.

\bibitem{He2016Deep}
K.~He, X.~Zhang, S.~Ren, and J.~Sun.
\newblock Deep residual learning for image recognition.
\newblock In {\em CVPR}, pages 770--778, 2016.

\bibitem{itti98}
L.~Itti, C.~Koch, and E.~Niebur.
\newblock A model of saliency-based visual attention for rapid scene analysis.
\newblock {\em IEEE TPAMI}, 20(11):1254--1259, 1998.

\bibitem{jia2014caffe}
Y.~Jia, E.~Shelhamer, J.~Donahue, S.~Karayev, J.~Long, R.~Girshick,
  S.~Guadarrama, and T.~Darrell.
\newblock Caffe: Convolutional architecture for fast feature embedding.
\newblock {\em arXiv preprint arXiv:1408.5093}, 2014.

\bibitem{jiang2013salient}
H.~Jiang, J.~Wang, Z.~Yuan, Y.~Wu, N.~Zheng, and S.~Li.
\newblock Salient object detection: A discriminative regional feature
  integration approach.
\newblock In {\em CVPR}, pages 2083--2090, 2013.

\bibitem{lee2016deep}
G.~Lee, Y.-W. Tai, and J.~Kim.
\newblock Deep saliency with encoded low level distance map and high level
  features.
\newblock In {\em CVPR}, pages 660--668, 2016.

\bibitem{li2015visual}
G.~Li and Y.~Yu.
\newblock Visual saliency based on multiscale deep features.
\newblock In {\em CVPR}, pages 5455--5463, 2015.

\bibitem{LiYu16}
G.~Li and Y.~Yu.
\newblock Deep contrast learning for salient object detection.
\newblock In {\em CVPR}, pages 478--487, 2016.

\bibitem{li2013saliency}
X.~Li, H.~Lu, L.~Zhang, X.~Ruan, and M.-H. Yang.
\newblock Saliency detection via dense and sparse reconstruction.
\newblock In {\em ICCV}, pages 2976--2983, 2013.

\bibitem{Li2016DeepSaliency}
X.~Li, L.~Zhao, L.~Wei, M.-H. Yang, F.~Wu, Y.~Zhuang, H.~Ling, and J.~Wang.
\newblock Deepsaliency: Multi-task deep neural network model for salient object
  detection.
\newblock {\em IEEE TIP}, 25(8):3919--3930, 2016.

\bibitem{li2016r}
Y.~Li, K.~He, J.~Sun, et~al.
\newblock R-fcn: Object detection via region-based fully convolutional
  networks.
\newblock In {\em NIPS}, pages 379--387, 2016.

\bibitem{li2014secrets}
Y.~Li, X.~Hou, C.~Koch, J.~Rehg, and A.~Yuille.
\newblock The secrets of salient object segmentation.
\newblock In {\em CVPR}, pages 280--287, 2014.

\bibitem{liu2016dhsnet}
N.~Liu and J.~Han.
\newblock Dhsnet: Deep hierarchical saliency network for salient object
  detection.
\newblock In {\em CVPR}, pages 678--686, 2016.

\bibitem{long2015fully}
J.~Long, E.~Shelhamer, and T.~Darrell.
\newblock Fully convolutional networks for semantic segmentation.
\newblock In {\em CVPR}, pages 3431--3440, 2015.

\bibitem{mahadevan2013biologically}
V.~Mahadevan and N.~Vasconcelos.
\newblock Biologically inspired object tracking using center-surround saliency
  mechanisms.
\newblock {\em IEEE TPAMI}, 35(3):541--554, 2013.

\bibitem{mahendran2015understanding}
A.~Mahendran and A.~Vedaldi.
\newblock Understanding deep image representations by inverting them.
\newblock In {\em CVPR}, pages 5188--5196, 2015.

\bibitem{noh2015learning}
H.~Noh, S.~Hong, and B.~Han.
\newblock Learning deconvolution network for semantic segmentation.
\newblock In {\em ICCV}, pages 1520--1528, 2015.

\bibitem{qin2015saliency}
Y.~Qin, H.~Lu, Y.~Xu, and H.~Wang.
\newblock Saliency detection via cellular automata.
\newblock In {\em CVPR}, pages 110--119, 2015.

\bibitem{ren2014region}
Z.~Ren, S.~Gao, L.-T. Chia, and I.~W.-H. Tsang.
\newblock Region-based saliency detection and its application in object
  recognition.
\newblock {\em IEEE TCSVT}, 24(5):769--779, 2014.

\bibitem{ronneberger2015u}
O.~Ronneberger, P.~Fischer, and T.~Brox.
\newblock U-net: Convolutional networks for biomedical image segmentation.
\newblock In {\em MICCAI}, pages 234--241, 2015.

\bibitem{siagian2007rapid}
C.~Siagian and L.~Itti.
\newblock Rapid biologically-inspired scene classification using features
  shared with visual attention.
\newblock {\em IEEE TPAMI}, 29(2):300--312, 2007.

\bibitem{simonyan2013deep}
K.~Simonyan, A.~Vedaldi, and A.~Zisserman.
\newblock Deep inside convolutional networks: Visualising image classification
  models and saliency maps.
\newblock {\em arXiv preprint arXiv:1312.6034}, 2013.

\bibitem{simonyan2014very}
K.~Simonyan and A.~Zisserman.
\newblock Very deep convolutional networks for large-scale image recognition.
\newblock {\em arXiv preprint arXiv:1409.1556}, 2014.

\bibitem{sun2011scale}
J.~Sun and H.~Ling.
\newblock Scale and object aware image retargeting for thumbnail browsing.
\newblock In {\em ICCV}, pages 1511--1518, 2011.

\bibitem{tong2015bootstrap}
N.~Tong, H.~Lu, X.~Ruan, and M.-H. Yang.
\newblock Salient object detection via bootstrap learning.
\newblock In {\em CVPR}, pages 1884--1892, 2015.

\bibitem{Vinyals2014Show}
O.~Vinyals, A.~Toshev, S.~Bengio, and D.~Erhan.
\newblock Show and tell: A neural image caption generator.
\newblock In {\em CVPR}, pages 3156--3164, 2014.

\bibitem{wang2015deep}
L.~Wang, H.~Lu, X.~Ruan, and M.-H. Yang.
\newblock Deep networks for saliency detection via local estimation and global
  search.
\newblock In {\em CVPR}, pages 3183--3192, 2015.

\bibitem{Wang2017CVPR}
L.~Wang, H.~Lu, Y.~Wang, M.~Feng, D.~Wang, B.~Yin, and X.~Ruan.
\newblock Learning to detect salient objects with image-level supervision.
\newblock In {\em CVPR}, 2017.

\bibitem{Wang2015Visual}
L.~Wang, W.~Ouyang, X.~Wang, and H.~Lu.
\newblock Visual tracking with fully convolutional networks.
\newblock In {\em ICCV}, pages 3119--3127, 2015.

\bibitem{wang2016saliency}
L.~Wang, L.~Wang, H.~Lu, P.~Zhang, and X.~Ruan.
\newblock Saliency detection with recurrent fully convolutional networks.
\newblock In {\em ECCV}, pages 825--841, 2016.

\bibitem{xie2015holistically}
S.~Xie and Z.~Tu.
\newblock Holistically-nested edge detection.
\newblock In {\em ICCV}, pages 1395--1403, 2015.

\bibitem{yan2013hierarchical}
Q.~Yan, L.~Xu, J.~Shi, and J.~Jia.
\newblock Hierarchical saliency detection.
\newblock In {\em CVPR}, pages 1155--1162, 2013.

\bibitem{yang2013saliency}
C.~Yang, L.~Zhang, H.~Lu, X.~Ruan, and M.-H. Yang.
\newblock Saliency detection via graph-based manifold ranking.
\newblock In {\em CVPR}, pages 3166--3173, 2013.

\bibitem{yang2016end}
W.~Yang, W.~Ouyang, H.~Li, and X.~Wang.
\newblock End-to-end learning of deformable mixture of parts and deep
  convolutional neural networks for human pose estimation.
\newblock In {\em CVPR}, pages 3073--3082, 2016.

\bibitem{zeiler2014visualizing}
M.~D. Zeiler and R.~Fergus.
\newblock Visualizing and understanding convolutional networks.
\newblock In {\em ECCV}, pages 818--833, 2014.

\bibitem{zhao2015saliency}
R.~Zhao, W.~Ouyang, H.~Li, and X.~Wang.
\newblock Saliency detection by multi-context deep learning.
\newblock In {\em CVPR}, pages 1265--1274, 2015.

\end{thebibliography}
}
\end{document}